\documentclass[lettersize,journal]{IEEEtran}
\usepackage[ruled,vlined]{algorithm2e}
\usepackage{amsmath,amsfonts,amssymb}

\usepackage{array}
\usepackage[caption=false,font=normalsize,labelfont=sf,textfont=sf]{subfig}
\usepackage{textcomp}
\usepackage{stfloats}
\usepackage{url}
\usepackage{verbatim}
\usepackage{graphicx}
\usepackage{cite}
\usepackage{booktabs}   % 三线表
\usepackage{pifont} %打叉打勾
\usepackage{threeparttable} %表格脚注
\usepackage{multirow}        % multi-row cells

\newcommand{\RNum}[1]{\uppercase\expandafter{\romannumeral #1\relax}}
\begin{document}

\title{PASTE: Physics-Aware Scattering Topology Embedding Framework for SAR Object Detection}

\author{Jiacheng Chen,~\IEEEmembership{Student Member,~IEEE}, Yuxuan Xiong, ~\IEEEmembership{Student Member,~IEEE}, Haipeng Wang,~\IEEEmembership{Senior Member,~IEEE}
        % <-this % stops a space
\thanks{The authors are with the Key Laboratory for Information Science of Electromagnetic Waves (MoE), Fudan University, Shanghai 200433, China (e-mail: jcchen21@m.fudan.edu.cn; 24210720295@m.fudan.edu.cn; 25113090103@m.fudan.edu.cn; hpwang@fudan.edu.cn}% <-this % stops a space

\thanks{This work was supported by the National Natural Science Foundation of China under Grant 62271153.}

}

\markboth{Journal of \LaTeX\ Class Files,~Vol.~14, No.~8, August~2021}%
{Shell \MakeLowercase{\textit{et al.}}: A Sample Article Using IEEEtran.cls for IEEE Journals}


\maketitle

\begin{abstract}
Current deep learning-based object detection paradigms for Synthetic Aperture Radar (SAR) imagery largely rely on transferring optical methodologies, often treating targets as texture patches while neglecting the intrinsic electromagnetic scattering mechanisms. Although scattering points have been explored in several studies to enhance detection performance, most existing methods still rely on amplitude-based statistical formulations. Alternative approaches have attempted to incorporate frequency-domain information for scattering center extraction; they generally suffer from heavy computational overhead and poor compatibility with diverse detection datasets. Consequently, these methods face significant challenges in embedding scattering topological information into contemporary detection frameworks effectively. To overcome the above challenges, this article proposes the Physics-Aware Scattering Topology Embedding Framework (PASTE), a novel closed-loop architecture designed for holistic scattering prior integration. By constructing the entire process from topological generation and injection to joint supervision, PASTE provides an elegant and efficient solution for incorporating scattering physics into modern SAR detectors. Specifically, it introduces a scattering keypoint generation and automated annotation scheme based on the Attributed Scattering Center (ASC) model to produce scalable and physics-consistent priors. Then, a scattering topology injection module is designed to guide multi-scale feature learning, while a corresponding scattering prior supervision strategy softly constrains network optimization by aligning predictions with the physical distribution of scattering centers. Extensive experiments on the realistic dataset demonstrate that PASTE is compatible with diverse detector architectures while yielding relative mAP gains of 2.9\% to 11.3\% across various baselines with acceptable computational overhead. Visual analysis of the predicted scattering map further demonstrates that PASTE successfully incorporates scattering topological priors into the feature space, exhibiting the ability to clearly distinguish target and background scattering regions, thus providing strong interpretability for the experimental results.
\end{abstract}

\begin{IEEEkeywords}
Synthetic Aperture Radar (SAR), Physics-Explainable Target Detection, Attribute Scattering Center Model, Electromagnetic Scattering Topology, Physics Prior Fusion
\end{IEEEkeywords}

\begin{figure}
    \centering
    \includegraphics[width=1\linewidth]{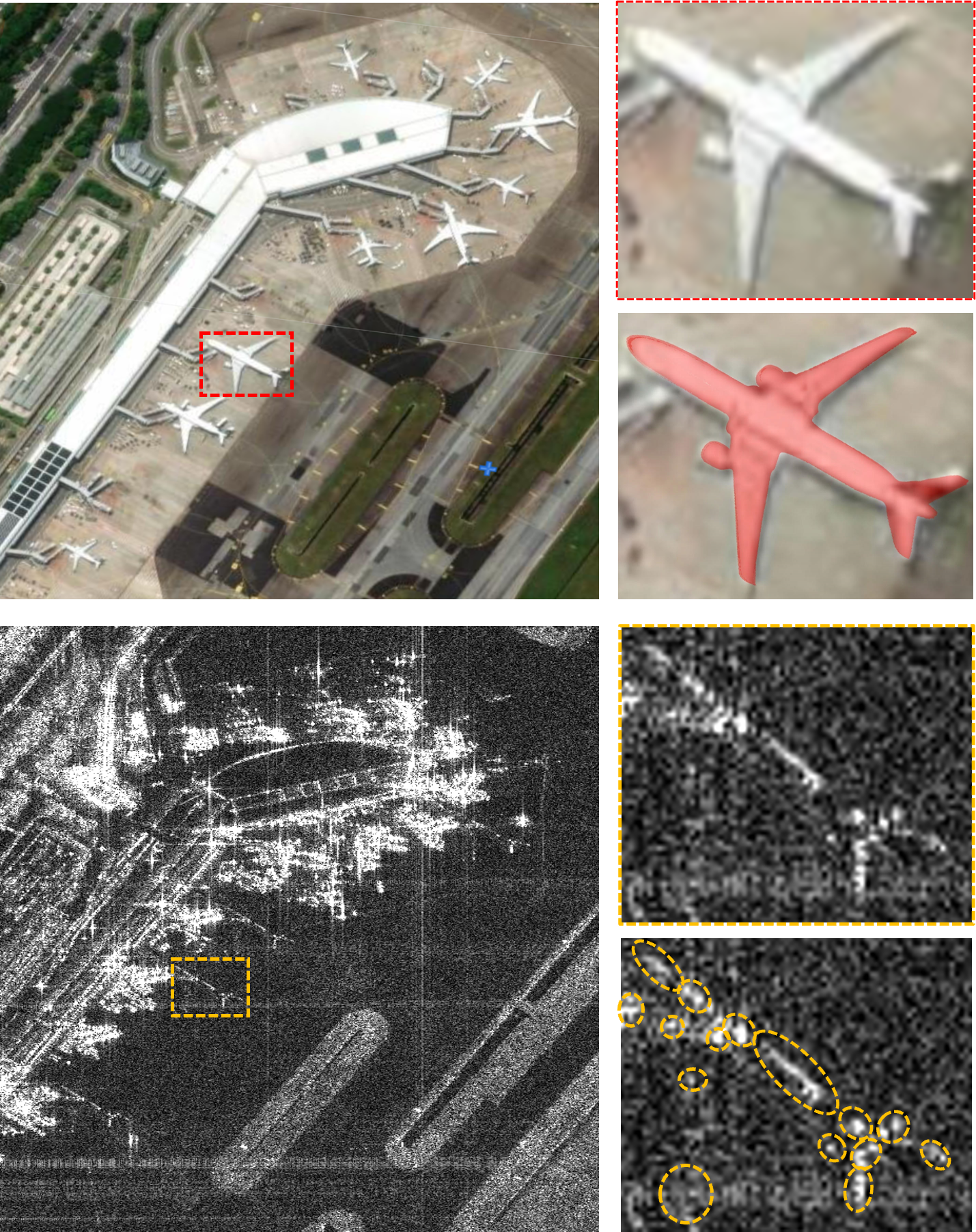}
    \caption{Optical (top) and SAR (bottom) images of the same terminal area in Changi Airport. The right side of the figure highlights the same aircraft target using a red region and yellow dashed lines, respectively. It can be observed that the aircraft target appears as a continuous region in the optical image, whereas it is characterized by discrete bright patches in the SAR image.}
    \label{fig:optical-SAR}
\end{figure}

\section{Introduction}
\IEEEPARstart{W}{ith} the rapid proliferation of Synthetic Aperture Radar (SAR) data and the continuous advancement of deep learning technologies, object detection and recognition in massive SAR imagery have become increasingly reliant on Deep Learning (DL)-based frameworks. DL–based detection and recognition frameworks have already achieved remarkable success and matured in the domain of optical imagery \cite{survey1, survey2, zhang2023remote}. Thus, many existing approaches for SAR image interpretation are directly adapted from the optical domain \cite{HUGHES2020166}. However, it brings some limitations since there are significant differences between SAR and optical images. As shown in Fig. \ref{fig:optical-SAR}, for the same type of object, optical images typically exhibit continuous geometric boundaries and coherent texture regions. In contrast, objects in SAR images appear as a series of discrete scattering centers \cite{li2021multiscale, li2019sar, Hu2024Conceptual}, resulting in a representation that is far less intuitive or directly interpretable. The geometric shape of the target can only be inferred after integrating these sparse scattering centers across the entire object \cite{Guo2020ExplainableSAR}. This fundamental difference makes the direct transfer of optical detection and recognition paradigms to SAR prone to limitations in accuracy, robustness, and interpretability \cite{HZL2020, feng2022electromagnetic, zhang2020fec, GNN-FiLM, zhang2021integrating, liu2021eftl}. Therefore, this work aims to introduce a fusion method that integrates physically meaningful scattering information into the existing detectors, thereby improving their performance on SAR imagery.

Most existing SAR detection methods are directly adapted from optical frameworks, spanning two-stage (e.g., Faster/Cascade R-CNN \cite{ren2015faster, cai2018cascade}) and one-stage series—including both anchor-based extensions (e.g., Oriented R-CNN \cite{xie2021oriented}, R3Det \cite{yang2021r3det}, S2A-Net \cite{han2021align}), and anchor-free designs (e.g., FCOS \cite{tian2019fcos}, CenterNet \cite{duan2019centernet}, YOLO series \cite{redmon2016you, cai2024yolov8})—as well as more recent Transformer-based variants \cite{carion2020end, 2024OEGR}. These methods share an implicit assumption: target regions can be treated as statistical texture patches, with their separability mainly determined by intensity and local gradient patterns aggregated through dense convolutions or self-attention. In essence, they learn target regions as local texture or intensity distributions. While this approach is feasible, it overlooks the distinctions between SAR and optical imagery: SAR echoes are determined by a limited number of scatterers and their geometric configurations, resulting in sparse, bright scattering centers overlaid on speckle backgrounds. Moreover, SAR target datasets often suffer from small-sample issues, where methods that rely purely on large-scale data without incorporating any electromagnetic prior knowledge are prone to underfitting or overfitting \cite{yin2024review}. Therefore, some studies have attempted to incorporate information such as amplitude, phase, polarization, or complex representations into SAR target detection fusion algorithms to address the aforementioned challenges \cite{zhou2024simulated}. Most existing fusion methods adopt channel concatenation or early fusion, directly stacking amplitude with phase (or real/imaginary components) and polarization-decomposed features as inputs to convolutional networks \cite{wang2021multichannel}; or they employ channel and spatial attention mechanisms such as Squeeze-and-Excitation \cite{makhija2024polsar}, multi-branch late fusion strategies \cite{shi2025multi}, Transformer/Cross-Attention fusion \cite{imani2025attention}, as well as complex/polarization-specific CNNs combined with scattering decomposition features \cite{wang2023target}. However, these fusion strategies typically overlook the spatial discontinuity of SAR targets, which consist of discrete scattering hotpots. Treating these disjointed spots as continuous entities can inevitably introduce cluttered background scattering, leading the model to struggle with missed detections of elongated targets and frequent false alarms caused by background pseudo-bright spots.

To integrate physical prior into SAR detection framework, its generation and representation are also core issues since there has long been a lack of scalable scattering-center–level structural annotations. Most existing datasets provide only bounding-box or rotated-box labels, with only a few recent complex-valued datasets (e.g., FAIR-CSAR-V1.0 \cite{wu2024fair}) including keypoint annotations derived from image intensity. However, scattering-center–level annotations for SAR imagery are not scalable when relying on traditional scattering center extraction algorithms or manual fine labeling, due to limitations in cost, stability, and format compatibility. This directly hinders the integration of scattering topology and physical attributes into DL.

Facing the aforementioned challenges, this paper proposes a novel fusion framework that effectively tackles the generation of scattering priors during the training of diverse existing detectors. Furthermore, dedicated scattering prior injection mechanisms and supervision strategies are developed to enhance target representation. The main contributions of this paper can be summarized as follows: 

\begin{itemize}
    \item [1)] This paper proposes a novel fusion framework that bridges the gap between physical scattering mechanisms and DL-based SAR detectors, named Physics-Aware Scattering Topology Embedding (PASTE) framework. By constructing a comprehensive closed-loop architecture of "generation, injection, and supervision". PASTE achieves a seamless integration of physical scattering prior into existing detection frameworks, effectively guiding the model to focus on the target's intrinsic electromagnetic characteristics rather than redundant background noise.
    \item [2)] An ASC-based scattering keypoint generation method and its automatic annotation process, named Scattering Keypoint Automatic Annotation (SKAA), is designed for the "Generation". By extracting ASC positional parameters, the scattering center is explicitly encoded as sparse keypoints to form the target scattering topology. This automated process decouples scattering centers in the image domain and fits ASC model parameters, using clustering to generate a fixed number of keypoints for each target, making zero manual annotation possible.
    \item [3)] A scattering map prediction network capable of learning the spatial distribution of target scattering keypoints with a supervision strategy responsible for constraining it during training are jointly proposed for the "Injection" and "Supervision", namely Scattering Topology Injection Module (STIM) and Scattering Prior Supervision Strategy (SPSS), respectively. By injecting physics-driven structural cues into the Feature Pyramid Network (FPN), the detector is guided to prioritize regions of scattering keypoints over all features. Correspondingly, SPSS employs keypoint-centered soft supervision as a physics-inspired constraint to align network attention with target scattering centers, enabling scattering prior supervision.
    \item [4)] Extensive experiments on the benchmarks with different imaging modes demonstrate that the proposed PASTE is architecture-agnostic and can be seamlessly integrated into diverse mainstream detectors, consistently yielding significant performance gains. Moreover, ablation studies verify that ASC-based scattering keypoints provide a more robust and discriminative structural prior than amplitude-based keypoints, underscoring the effectiveness of physics-driven scattering topology for SAR target detection.
\end{itemize}

The remainder of this paper is organized as follows. Section \RNum{2} reviews related work on SAR object detection and methods that incorporate physical scattering priors. Section \RNum{3} details our proposed method, beginning with the overall architecture. We then elaborate on our three core contributions. Section \RNum{4} describes our experimental setup and training scheme details, and presents comparison experiments and ablation studies to validate the effectiveness of our proposed components. Subsequently, we demonstrate the superiority of our approach through quantitative comparisons with baseline methods and qualitative visualizations. Section \RNum{5} concludes the paper by summarizing our main contributions and findings. Finally, we explore potential directions for future research.

\section{Related Work}

This section reviews related work on SAR object detection and recognition algorithms. It begins with traditional SAR object detection methods, followed by the introduction of DL-based approaches. Subsequently, we discuss methods that fuse SAR-specific characteristics for improved detection and recognition. Finally, we cover several advanced fusion methods that leverage keypoint information for enhancement.

\subsection{Traditional Methods for SAR Object Detection}
Traditional methods for SAR target detection often rely on image processing and statistical approaches, encompassing techniques such as the Hough transform for edge detection, Support Vector Machines (SVM), and the Constant False Alarm Rate (CFAR) detector \cite{ozkaya2020automatic, shao2023CFARG}. Among these, CFAR is particularly representative due to its adaptive thresholding capability and its effectiveness in homogeneous backgrounds. The core principle of the CFAR detector is to establish a dynamic local threshold. It employs a sliding window to analyze the background region surrounding a Cell Under Test (CUT) and estimates the statistical distribution of the local clutter (e.g., Rayleigh, Weibull, or K-distribution). Based on a predefined false alarm rate, an adaptive detection threshold is then computed. Despite the effectiveness of CFAR and its variants under specific conditions, their limitations are significant in ignoring rich semantic information such as shape, texture, structure, and context. This reliance on a single feature often leads to a high rate of false alarms in complex backgrounds.

Subsequently, researchers proposed more advanced methods to overcome these challenges. Addressing the impact of non-uniform backgrounds on detection robustness in high-resolution SAR images, Hu et al. \cite{hu2019aircraft} developed a detection method based on a Generalized Gamma Mixture Distribution and a mixture model. Compared to basic CFAR, this approach effectively reduces the false alarm rate in complex backgrounds. In another line of work, Ding et al. \cite{ding2017robust} introduced the electromagnetic scattering representational models. They modeled individual ASC \cite{potter1997attributed} using an adaptive Gaussian distribution and employed Kullback-Leibler (KL) divergence to measure the similarity between the image and templates, thereby accomplishing the detection and recognition task.

\subsection{DL-Based Methods for SAR Object Detection}
Given the limitations of traditional methods, which rely on manual feature engineering and exhibit poor generalization, the research community has increasingly turned its attention to DL. Pioneering work by Chen et al. \cite{chen2014CNN} applied Fully Convolutional Networks (FCN) to SAR target recognition, demonstrating the significant potential of DL models for processing SAR data and achieving high accuracy. To address the need for real-time processing, Cui et al. \cite{TNN_cui} proposed a lightweight, threshold-based neural network (TNN) that predicts an optimal detection threshold within a sliding window, drastically increasing the detection speed (FPS). Another study by Chen et al. \cite{chen2024yolo} improved the YOLO network architecture by integrating an attention mechanism to fuse multi-scale features, thereby mitigating issues with overlapping and missed detections. The powerful feature extraction capabilities of the Transformer architecture \cite{NIPS2017_Transformer}, following its migration from Natural Language Processing (NLP) to the Computer Vision (CV) domain, have also captured the interest of researchers. For instance, Chen et al. \cite{chen2022geospatial} developed a Geospatial Transformer framework that enhances multi-scale feature extraction and effectively suppresses background noise by introducing a Pyramid Convolutional Attention fusion module and a Residual Spatial Attention module. Building on the Transformer architecture, Jia et al. \cite{Jia2024Ship} achieved precise detection and identification of ships in ports and open seas by supplementing the model with accurate sea-land segmentation and false alarm filtering modules, delivering state-of-the-art performance on the FuSAR-Ship \cite{fusarShip2020} dataset. More recently, Large Language Models (LLMs) have achieved remarkable success in optical vision tasks, benefiting from large-scale multimodal pretraining and strong semantic alignment capabilities \cite{10.1093/nsr/nwae403, MA2026VLM}. However, their powerful generalization ability is largely derived from rich multimodal semantic priors inherent in optical images, whereas SAR targets are predominantly composed of sparse and discrete scattering centers, which are difficult to be effectively captured by language-aligned representations.

Some studies have brought the concept of structural anchor points in this field. Recognizing that strong scattering points on SAR targets are often sparsely distributed, Kang et al. \cite{Kang2022SFR} proposed a scattering point relation module to model the correlations between them. Along similar lines, Meng et al. \cite{Meng2025STC} subsequently introduced STC-Net, which employs a Star-shaped Topology (ST) derived from scattering cues to model the target and highlight its key components. To address the challenge of open-set recognition, Xiao et al. \cite{xiao2025RPL} utilized Reciprocal Point (RPL) to construct a paradigmatic feature that defines a bounded space for known classes in the feature domain. However, a common thread among these methods is that while they are all conceptually based on structural anchors, their keypoints are typically derived from image processing techniques and lack intrinsic electromagnetic physical properties.

\subsection{DL-based Methods Fused with SAR Scattering Characteristics}

Although the aforementioned DL-based methods have achieved significant progress, they largely treat SAR images as generic 2D images, overlooking their unique electromagnetic scattering physics, which limits the models' performance and interpretability in complex scenarios. Thus, several recent studies have exploited scattering characteristics to improve SAR target detection performance. 

\begin{figure*}
    \centering
    \includegraphics[width=1\linewidth]{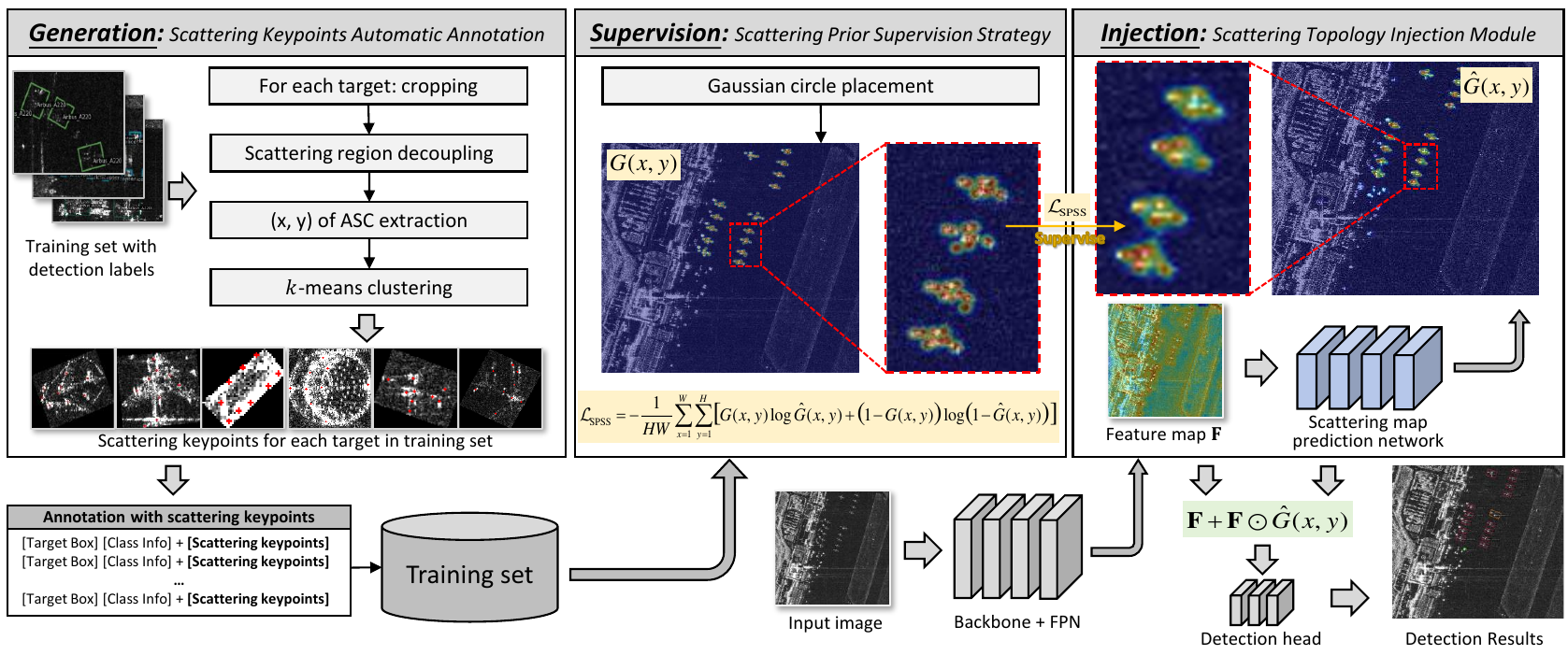}
    \caption{The overview architecture of the proposed PASTE that consists of three components: Generation, Supervision, and Injection.
During training, the Scattering Keypoints Automatic Annotation (SKAA) for "Generation" automatically generates scattering keypoints and forms extended annotation files that are fully compatible with the standard detection training pipeline.
Then Scattering Prior Supervision Strategy (SPSS) for “Supervision” aligns the generated keypoints at the image scale and supervises the scattering map prediction network within the Scattering Topology Injection Module (STIM), guiding it to learn the scattering topology prior and predict target scattering regions. During inference, STIM for "Injection" implicitly encodes the learned scattering topology prior and can directly predict the scattering map from the input image without the involvement of SKAA and SPSS, enabling feature fusion with existing detectors.}
    \label{fig:ASTFM}
\end{figure*}

Fu et al. \cite{fu2021scattering} enhanced target scattering regions using an improved Harris-Laplace detector with DBSCAN clustering and GMM-KLD discrimination, then fed the preprocessed image into an attention pyramid network incorporating multi-scale FPN, improved CBAM, and focal loss. Guo et al. \cite{guo2020scattering} proposed the anchor-free SKG-Net guided by scattering keypoints, employing context-aware feature selection and deformable convolutions for oriented ship detection. Sun et al. \cite{sun2022span} introduced SPAN for unified detection and classification, integrating scattering feature extraction, resolution recovery, and automatic RoI generation modules to model global scattering distribution while suppressing land interference. Wang et al. \cite{wang2025attributed} developed the ASC-guided ASC-U²Det with omnidirectional sub-aperture segmentation and a dedicated SAR-Vehicle-Det dataset, using reconstructed ASC masks to extract azimuth-dependent scattering features. Pan et al. \cite{pan2024sffnet} designed the dual-branch SFFNet with scattering center reconstruction and a dedicated attention fusion module combined with dense depthwise separable blocks for efficient maritime ship detection. Yang et al. \cite{YY2025SARDet} proposed SARDet-CL, a self-supervised contrastive learning framework that incorporates feature enhancement and imaging-mechanism constraints to generate more discriminative representations for SAR target detection, thereby significantly improving detection performance. Zhao et al. \cite{Zhao2024Azimuth} proposed an azimuth-aware subspace classifier on the Grassmann manifold to address the few-shot class-incremental learning problem in SAR automatic target recognition, demonstrating superior and stable performance. Huang et al. \cite{Huang2025PGD} propose a physics-guided learning paradigm for SAR airplane detection, which encodes prior knowledge of discrete structural distributions via physics-guided self-supervised learning, enhances multi-scale feature representations, and focuses on dominant scattering points at the detection head, thereby improving fine-grained detection and recognition performance in complex backgrounds. 

In contrast, this paper explores an ASC-based scattering topological fusion framework that integrates the generation, injection, and supervision of scattering priors. This framework effectively combines local scattering keypoint priors with the global contextual perception capability of DL networks, resulting in significant performance improvements.

\section{Proposed Method}
In the section, an overview of the proposed PASTE framework is first presented. Then, the three core components of PASTE are elaborated in detail, including a novel scattering keypoint identification and automatic annotation method, a scattering topology injection module for multi-scale feature learning, and a scattering prior supervision strategy that aligns network optimization with the physical distribution of scattering centers.

\subsection{Overall Architecture of PASTE}
The overall architecture of the proposed PASTE is illustrated in Fig. \ref{fig:ASTFM}, and consists of three main components: Scattering Keypoints Automatic Annotation (SKAA) method, the Scattering Topology Injection Module (STIM), and the corresponding Scattering Prior Supervision Strategy (SPSS).

The general workflow of PASTE is as follows. For SAR object detection datasets, SKAA is first applied to transform existing target-level rotated box annotations into scattering center-level annotations, thereby generating precise scattering keypoints. Subsequently, the STIM is inserted after the backbone network and FPN to capture and enhance the original pyramid features. Within STIM, a parallel keypoint distribution prediction branch produces spatial scattering maps with resolutions matching the shallow feature layers. The predicted spatial scattering maps are supervised via the SPSS, which utilizes scattering keypoint annotations from SKAA to construct Gaussian distributions at the corresponding spatial positions, thereby yielding a standard scattering topological map as supervisory signals. During training, STIM is jointly optimized by both SPSS and the downstream detection objectives, such as classification and regression loss. Together, PASTE forms a novel fusion paradigm of scattering priors and DL-based methods.  

\subsection{Scattering Keypoints and Automatic Annotation Method}

\begin{figure*}
    \centering
    \includegraphics[width=1\linewidth]{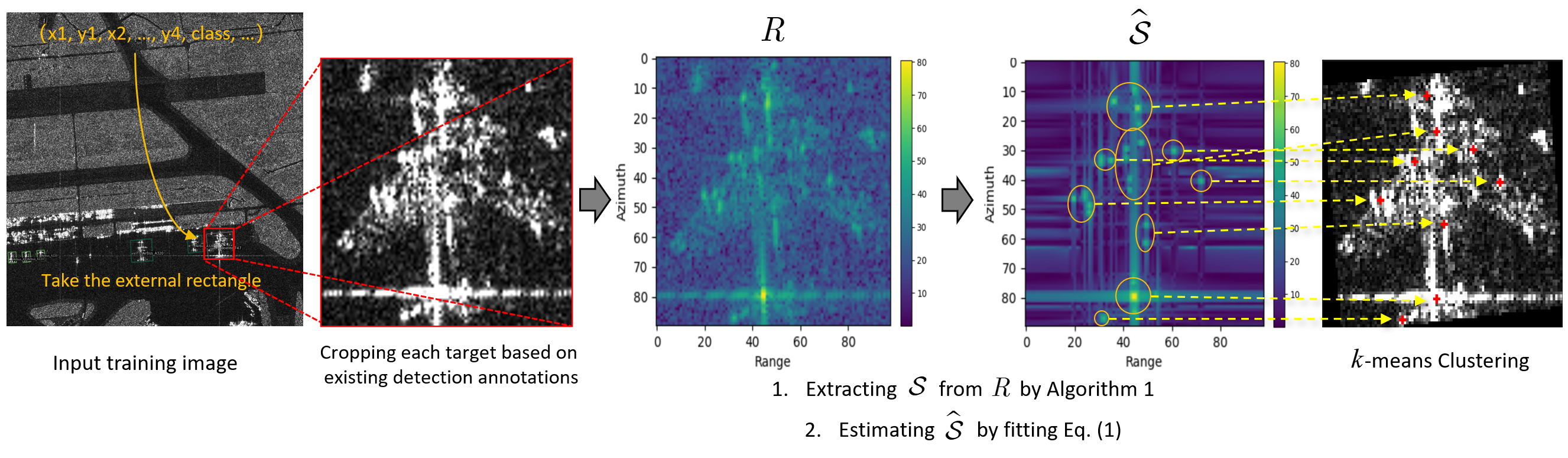}
    \caption{Workflow of the SKAA method}
    \label{fig:SKAA}
\end{figure*}

\begin{algorithm}[t]
\caption{Image-domain Scattering Region Decoupling}
\label{alg:decouple}

\KwIn{Complex image $I \in \mathbb{C}^{H \times W}$, dB threshold $\tau$}
\KwOut{Decoupled scattering region list $\mathcal{S}$}

$R \leftarrow |I|$, $\mathcal{S} \leftarrow \emptyset$, $n \leftarrow 0$\;

\While{$n < N_{\max}$}{

    \textbf{(1) Scattering Block Masking based on Breadth First Search (BFS)}\;
    $\tau_p \leftarrow \max(R)\cdot 10^{\tau/10}$\;
    $(x_0,y_0)\leftarrow \arg\max R$\;
    Initialize binary mask $B \leftarrow 0$\;
    Initialize queue $\mathcal{Q}\leftarrow\{(x_0,y_0)\}$, set $B(x_0,y_0)=1$\;

    \While{$\mathcal{Q}$ not empty}{
        Pop $(x,y)$ from $\mathcal{Q}$\;
        \ForEach{$4$-neighbor $(x',y')$ of $(x,y)$}{
            \If{$(x',y')$ out of bounds}{continue}
            \If{$B(x',y')=1$}{continue}
            \If{$R(x',y')\le \tau_p$}{continue}
            $B(x',y')\leftarrow 1$\;
            Push $(x',y')$ into $\mathcal{Q}$\;
        }
    }

    \textbf{(2) Log-amplitude guided region growing}\;
    $p \leftarrow \max(R)$\;
    $D(x,y)\leftarrow 10\log_{10}\!\big((R(x,y)+\epsilon)/p\big)$\;
    $\Omega\leftarrow\{(x,y)\mid D(x,y)>-20\}$ (sorted desc.)\;
    $T\leftarrow B$, $l\leftarrow 1$\;

    \ForEach{$(x,y)\in\Omega$}{
        \If{$T(x,y)\neq 0$}{continue}
        $\mathcal{N}\leftarrow$ labels of $8$-neighbors of $(x,y)$, remove $0$\;
        \eIf{$\mathcal{N}=\emptyset$ \textbf{and} $D(x,y)>\tau$}{
            $l\leftarrow l+1$, $T(x,y)\leftarrow l$
        }{
            $T(x,y)\leftarrow \min(\mathcal{N})$
        }
    }

    \textbf{(3) Region extraction and residual update}\;
    $S(x,y)\leftarrow R(x,y)$ if $T(x,y)=1$, else $0$\;
    $\mathcal{S}\leftarrow \mathcal{S}\cup\{S\}$\;
    $R\leftarrow \max(R-S,0)$\;
    $n\leftarrow n+1$\;
}

\Return{$\mathcal{S}$}

\textbf{Parameter setting:} $\tau=-3$ dB, $\epsilon=10^{-6}$, $N_{\max}=20$.

\end{algorithm}

The ASC model is widely employed in physical prior fusion methods for SAR Automatic Target Recognition (ATR) tasks, owing to its superior capacity for modeling target scattering characteristics. However, due to computational complexity and target sparsity, it is impractical to directly characterize all scattering components using the ASC model in detection scenarios. Therefore, we constrain the ASC parameter extraction to regions delineated by existing target-level annotations within the detection dataset. This strategy ensures that the extracted scattering information remains both relevant and computationally feasible for downstream tasks. 

The specific process of SKAA is shown in Fig. \ref{fig:SKAA}. Firstly, based on the existing target box annotation, the image is cropped to obtain the target slice. Then, an image domain decoupling method that is adapted from \cite{Chen2024RL_SAR}, as described in Algorithm~\ref{alg:decouple} is designed and applied to separate the scattering centers of each target, yielding a set of scattering centers' regions $\mathcal{S}$. 

Each region is then fitted using the ASC model to extract the corresponding position parameters $(x_i, y_i)$. Specifically, the other parameters of ASC are assumed to be fixed, and only the position parameters are optimized to minimize the distance between the reconstructed scattering center $\Hat{S_i}$ and $S_i$, see Eq.~\eqref{eq:ASC_fitting}. The parameters $(x_i, y_i)$ are regularized according to \cite{Zhang2009ASC}, ensuring direct correspondence with image pixel positions and alignment with existing annotations.

\begin{equation}
(x_i, y_i) =
\arg\min_{(x, y)}
\left\|
S_i - \hat{S}_i(x, y)
\right\|_2
\label{eq:ASC_fitting}
\end{equation}

where $S_i$ denotes the scattering feature of the $i$-th target, $\hat{S}_i(x,y)$ represents the scattering feature extracted at spatial location $(x,y)$, and $(x_i, y_i)$ is the estimated coordinate of the $i$-th ASC obtained by minimizing the $\ell_2$ distance between them. $\Hat{S_i}(x_i, y_i)$ in Eq.~\eqref{eq:ASC_fitting} is given by Eq.~\eqref{eq:ASC_single} and Eq.~\eqref{eq:IFFT_win}.

\begin{equation}
    \mathbf{E}_i(x_i, y_i) = C(f, \varphi) \cdot \exp\left( -\mathrm{j} \frac{4\pi f}{c} (x_i \cos\varphi + y_i \sin\varphi) \right )
    \label{eq:ASC_single}
\end{equation}

\begin{equation}
    \Hat{S_i}(x_i, y_i) = \mathrm{IFFT2D}(w(\mathbf{E}_i(x_i, y_i)))
    \label{eq:IFFT_win}
\end{equation}

Given that the remaining ASC parameters are held constant, the formulation of the individual scattering field $\mathbf{E}_i(x_i, y_i)$ is simplified through the use of a constant matrix $C(f, \varphi)$, in which $f$ and $\varphi$ represent the frequency and azimuth, respectively. $c$ in Eq.~\eqref{eq:ASC_single} represents light speed. $\mathrm{IFFT2D(\cdot)}$ denotes the two-dimensional inverse Fourier transform, applied together with a window function $w(\cdot)$ to obtain the reconstructed scattering field image. In this work, the Taylor window is employed because of its superior sidelobe suppression capability.

To establish a unified topological representation and generate scattering center-level annotations, these extracted keypoints are subsequently clustered. The clustering process enables the consolidation of adjacent or redundant scatterers, effectively reducing interference from cluttered scattering points near the target and highlighting the spatial distribution of the dominant scattering centers, thereby improving the robustness and clarity of spatial attention generation. Specifically, the $k$-means clustering algorithm is employed to group all scattering centers associated with each target into nine distinct scattering keypoints, as determined through experimental evaluation. This fixed number not only standardizes the annotation format, but also ensures consistency in representing targets of different scales. Finally, the coordinates of the scattering keypoints are transformed into the global coordinate system consistent with the bounding box annotations and incorporated into the original labels of all instances in the dataset for training.

\subsection{Scattering Topology Injection Module}

With the prior knowledge of target scattering topology established, the next step is to incorporate this information into the feature representation process. Therefore, the STIM is proposed to effectively highlight scattering-sensitive regions across multiple spatial scales. The underlying concept of the STIM is intuitive. It constructs a spatial scattering map to reinforce the feature responses in regions corresponding to the scattering keypoints, thereby enhancing the network’s sensitivity to scattering-aware spatial information. In addition, the STIM is designed as a plug-and-play module that receives multi-scale features from both the backbone and the FPN. It outputs enhanced multi-scale features to the subsequent network without altering the original feature pyramid structure. 

\begin{figure}
    \centering
    \includegraphics[width=1\linewidth]{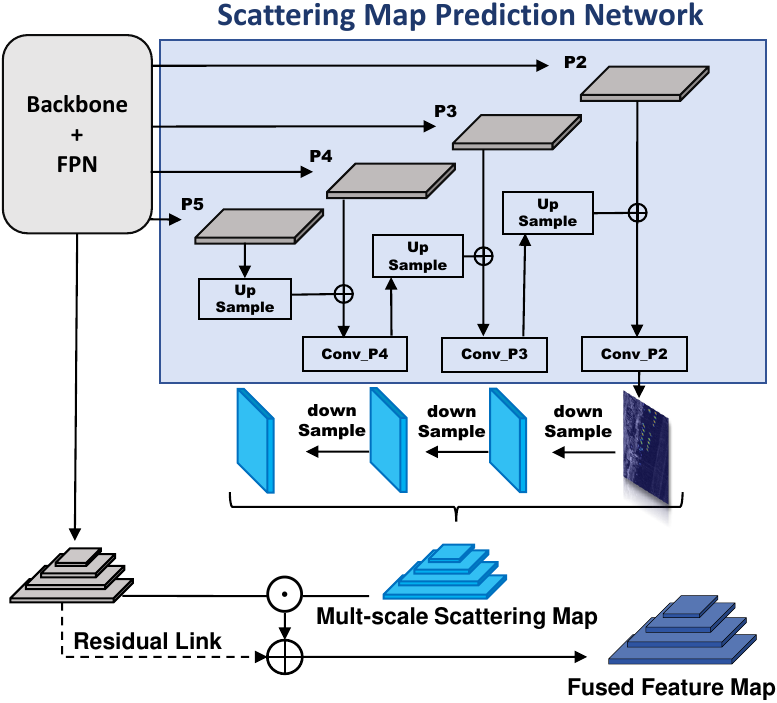}
    \caption{Diagram of spatial scattering map prediction network and fusion flow}
    \label{fig:SAMNet}
\end{figure}

The detailed architecture of scattering map prediction network in STIM and its fusion flow is illustrated in Fig. \ref{fig:SAMNet}. The design rationale of this network is as follows. Shallow feature maps contain richer spatial details, making them suitable for predicting the spatial distribution of target scattering area. However, high-level features carry stronger semantic information that is also beneficial for attention generation. Therefore, a progressive upsampling and concatenation strategy is adopted to integrate semantic cues from deep layers into the shallow ones, enabling the scattering map to jointly capture both spatial and semantic characteristics. High-level features (e.g., P5) are progressively upsampled and fused with lower-level FPN features (P4, P3, P2). Each step is followed by a 3×3 convolution (Conv\_P4, Conv\_P3, Conv\_P2) and ReLU activation. Finally, a 1×1 convolution and sigmoid function generate the spatial scattering map. To achieve efficient multi-scale scattering map generation, the scattering map is predicted only at the shallowest scale, where spatial details are most abundant. To prevent redundant computations from repeatedly predicting scattering map at each scale, scattering map for higher levels are produced through progressive downsampling as shown in Fig. \ref{fig:SAMNet}, thus completing the construction of the spatial attention pyramid.

Afterward, the predicted spatial scattering map is applied to the original FPN features via element-wise multiplication to emphasize scattering-relevant regions. The enhanced features are then added back to the original FPN outputs, forming the final enhanced feature $\mathbf{F}_p^{\mathrm{enh}}$. The process can be expressed as Eq.~\eqref{eq:Enhancment}:

\begin{equation}
\mathbf{F}_p^{\mathrm{enh}}=\mathbf{F}_p+\mathbf{A}_p \odot \mathbf{F}_p, \quad p \in\{P 2, P 3, P 4, P 5\}
\label{eq:Enhancment}
\end{equation}
where $\mathbf{F}_p$ denotes the original FPN feature at scale $p$, $\mathbf{A}_p$ represents the corresponding spatial scattering map, and $\odot$ indicates element-wise multiplication.

\subsection{Scattering Prior Supervision Strategy}

During inference, the SKAA method does not participate in the generation of scattering cues. Instead, the scattering topology enhancement is solely achieved by the scattering map prediction network, which has already learned scattering topology priors during training. This design ensures that the detection network equipped with PASTE retains its end-to-end inference property, without introducing any additional dependencies or decoupled processes. To enable this network to effectively capture the distribution of scattering keypoints while maintaining compatibility with the downstream detection objectives, a joint optimization strategy is adopted. Specifically, the proposed SPSS is combined with the detection loss to supervise the training of the scattering map prediction network. 

During training, the SPSS constructs a ground-truth scattering map to guide the predicted spatial scattering map toward the scattering keypoint locations. Ideally, the predicted attention distribution should exhibit high responses only around the scattering keypoints generated by the SKAA method. Thus, SPSS begins by initializing a zero-valued scattering map with the same spatial size as the input feature. A 2D Gaussian distribution is then placed at each annotated scattering keypoint position to model its spatial influence. When multiple scattering keypoints are spatially close and their Gaussian regions overlap, SPSS merges them using a pixel-wise maximum operation, ensuring that the strongest response is preserved in the overlapping areas. Formally, the construction of the ground-truth scattering map can be defined as follows. 

Let the spatial size of the input feature map be $H\times W$, the scattering keypoints set $\mathcal{K}$ is:

\begin{equation}
    \mathcal{K} = \{(x_k, y_k)\}_{k=1}^{N},
    \label{eq:SKPSet}
\end{equation}
where $x_k$, $y_k$ indicate the pixel coordinates of the corresponding keypoints. Then the ground-truth Gaussian scattering map $G(x, y)$ can be expressed as Eq.~\eqref{eq:GT map}.

\begin{align}
G(x, y) 
&= \max_{k=1,\dots,N} 
\exp\!\left(
    -\frac{(x - x_k)^2 + (y - y_k)^2}{2\sigma^2}
\right), \nonumber\\
&\quad x \in [0, W), \, y \in [0, H)
\label{eq:GT map}
\end{align}
where \(W\) and \(H\) represent the spatial extents in the horizontal and vertical directions, respectively. The set \(\{(x_k, y_k)\}_{k=1}^{N}\) corresponds to \(N\) given by scattering keypoints $\mathcal{K}$. The parameter \(\sigma > 0\) denotes the standard deviation of the Gaussian kernel, which controls the spatial spread of the response. The operator \(\max_{k=1,\dots,N}(\cdot)\) selects the maximum response over all reference points, such that the value of \(G(x,y)\) at each location is dominated by the scattering keypoints yielding the strongest Gaussian response. The resulting ground-truth scattering map thus captures the spatial topology of scattering keypoints while providing a smooth and continuous supervision signal for network learning. Accordingly, the learning objective of the attention prediction branch is defined using a Binary Cross-Entropy (BCE) loss as formulated in Eq.~\eqref{eq:SGA Loss}, which measures the pixel-wise difference between the predicted scattering map $\hat{G}(x,y)$ and ground-truth scattering map $G(x,y)$.

\begin{align}
    \mathcal{L}_{\mathrm{BCE}} = &
- \frac{1}{H W} 
\sum_{x=1}^{W} \sum_{y=1}^{H} 
\Big[
G(x, y) \log \hat{G}(x, y) + \notag \\
&(1 - G(x, y)) \log (1 - \hat{G}(x, y))
\Big],
\label{eq:SGA Loss}
\end{align}

In the training process, this loss only supervises the scattering map prediction branch, while the backbone and FPN weights remain unaffected.

\section{Experimental Results}

This section presents a series of experiments, results, and analyses to thoroughly demonstrate the effectiveness of the proposed PASTE. We first introduce the dataset used in the experiments. Next, we describe the detailed training procedure and showcase the scattering keypoint annotations generated by SKAA along with their visualizations. Then, we explain the performance evaluation metrics and their computation, followed by a discussion of the selected baseline detection algorithms and comparative experiments with and without PASTE. Next, a visual analysis of the predicted spatial scattering map is conducted. Moreover, ablation studies are conducted on the scattering keypoints and the loss function in SPSS. Finally, the additional spatiotemporal overhead introduced by PASTE is evaluated.

\begin{figure*}
    \centering
    \includegraphics[width=1\linewidth]{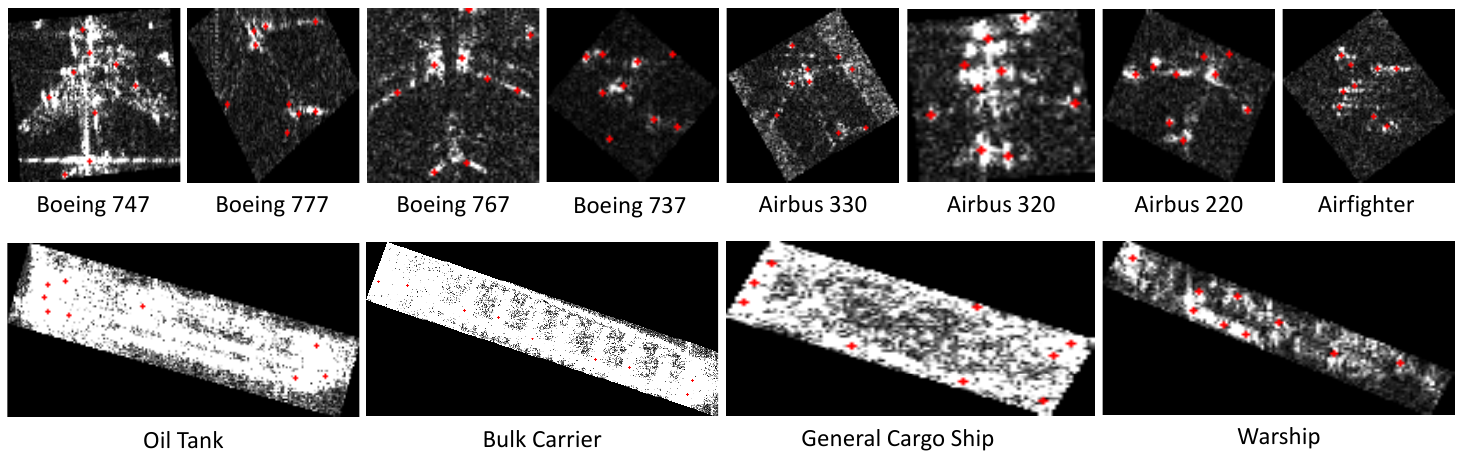}
    \caption{Visualization of scattering keypoints for representative classes in the FAIR-CSAR training set. The first row shows examples from major aircraft categories, while the second row illustrates keypoint patterns of representative ship categories. All automatically generated scattering keypoints are marked with red dots in the figure.}
    \label{fig: keypoints visualization}
\end{figure*}

\subsection{Datasets}

The FAIR-CSAR \cite{wu2024fair} dataset is adopted for the conduct of experiments, which is a fine-grained object detection and recognition benchmark built upon Single-Look Complex (SLC) SAR imagery. It aims to address several limitations of existing SAR datasets, including the absence of signal-domain information, coarse annotation granularity, and insufficient sample volume. FAIR-CSAR is constructed from 175 SLC images acquired by the Gaofen-3 satellite, covering 32 cities worldwide as well as multiple maritime regions. The dataset contains 349,002 annotated instances with a spatial resolution ranging from 1–5 m, and employs oriented bounding boxes (OBB) to support robust object analysis in complex scenes.

FAIR-CSAR consists of two imaging modes: Spotlight (SL) and Fine Strip-Map (FSI). 

\subsubsection{SL mode}
The SL mode constitutes the primary high-resolution acquisition protocol of the FAIR-CSAR dataset and is specifically designed to enable fine-grained object analysis in complex scenes through meticulous imaging. This mode delivers a nominal spatial resolution of 1 m and operates in single polarization, embracing all four polarimetric combinations (HH, HV, VH, VV). The current release encompasses 170k annotated instances distributed across 22 fine-grained categories, namely Airbus\_A220, Airbus\_A320, Airbus\_A330, Airfreighter, Boeing\_737, Boeing\_747, Boeing\_767, Boeing\_777, Fokker-50, Gulfstream, Helicopter, Other\_Aircraft, Bulk\_Carrier, Container\_Ship, General\_Cargo\_Ship, Motion-Defocusing\_Ship, Oil\_Tanker, Warship, Other\_Ship, Bridge, Tank, and Tower\_Crane.

\subsubsection{FSI mode}
The FSI mode is tailored for large-area surveillance at medium resolution. Acquired with a nominal ground-range resolution of 5 m and dual-polarimetric channels, this mode contains 170k labeled instances condensed into three major categories: Bridge, Ship, and Tank. FSI imagery spans oceans, rivers, and harbors, and faithfully reproduces real-world SAR phenomena such as motion defocusing and azimuth ambiguities.

Together, the SL and FSI modes partition the FAIR-CSAR dataset into two complementary subsets that span resolutions, polarizations, scene complexities, and category granularities, thereby providing a comprehensive and sufficiently challenging benchmark for robust experimental evaluation.

\subsection{Training Details}

Models are trained for 15 epochs on FAIR-CSAR SL mode and FSI mode training sets, respectively, and the checkpoint achieving the best performance during training is selected for final evaluation. The batch size is fixed to 4 for all configurations. For detectors based on ResNet-50, we use an SGD optimizer with an initial learning rate of 0.005, momentum of 0.9, and weight decay of $1\times 10^{-4}$, together with gradient clipping ($\mathrm{norm}_{\mathrm{max}}=35$). The learning rate follows a step decay schedule with linear warm-up for the first 500 iterations and decay points at epochs 8 and 11. For Swin-T–based detectors, AdamW is adopted with a learning rate of $1\times 10^{-4}$, weight decay of 0.05, and customized parameter-wise decay settings for positional embeddings and normalization layers. All experiments are conducted on a Linux server equipped with two NVIDIA RTX 2080 Ti GPUs. 

Before training detectors equipped with the proposed PASTE, the training split of each dataset is first pre-processed by the SKAA process to automatically generate scattering keypoint annotations. This procedure is fully automated and produces all required supervision signals before the training stage begins. Fig. \ref{fig: keypoints visualization} presents the visualized scattering keypoints generated by the SKAA module for some categories in the training set from FAIR-CSAR.  

\begin{table*}[t]
\centering
\caption{Baseline Detectors Adopted for PASTE Evaluation}
\label{tab:baselines}
\small
\renewcommand{\arraystretch}{1.3}          % 行距拉大 30 %
\begin{tabular}{@{}
>{\raggedright\arraybackslash}m{0.22\linewidth}
>{\centering\arraybackslash}m{0.06\linewidth}
>{\centering\arraybackslash}m{0.1\linewidth}
>{\raggedright\arraybackslash}m{0.55\linewidth}@{}}
\toprule
\textbf{Baseline Detector} & \textbf{Stage} & \textbf{Anchor} & \textbf{Justification for Selection}\\
\midrule
Rotated RetinaNet & single & anchor-based & Classic one-stage anchor-based baseline; widely used in rotated object detection.\\
GWD & single & anchor-free & Represents anchor-free regression with probabilistic localization loss.\\
RoI Transformer & two & anchor-based & Converts horizontal RoIs to rotated ones via learnable transformation; tests PASTE on geometry-aware proposal refinement.\\
Rotated FCOS & single & anchor-free & Dense anchor-free detector; tests PASTE on center-ness assignment.\\
Oriented R-CNN (ResNet-50) & two & anchor-based & Strong two-stage CNN baseline; demonstrates PASTE compatibility with standard backbones.\\
Oriented R-CNN (Swin-T) & two & anchor-based & Swin-Transformer variant; evaluates PASTE on Transformer backbones.\\
\bottomrule
\end{tabular}
\end{table*}

\subsection{Evaluation Metrics and Computation}

Model performance is evaluated using the mean Average Precision (mAP), the standard metric for object detection. For each class $c$, the Average Precision (AP) is computed as the area under the precision–recall curve:

\begin{equation}
\mathrm{AP}_c=\int_0^1 p_c(r) d r
\label{eq: AP}
\end{equation}
where $p_{c}(r)$ denotes the precision at recall level $r$. The mean Average Precision is obtained by averaging Eq.~\eqref{eq: AP} over $C$ classes:

\begin{equation}
\mathrm{mAP}=\frac{1}{C} \sum_{c=1}^C \mathrm{AP}_c
\label{eq: mAP}
\end{equation}

In our experiments, we adopt mAP@0.5, where a predicted oriented bounding box $\hat{B}$ is considered correct if its Intersection-over-Union (IoU) with the ground truth box $B$ satisfies:

\begin{equation}
\operatorname{IoU}(B, \hat{B})=\frac{|B \cap \hat{B}|}{|B \cup \hat{B}|}>0.5
\label{eq: IOU}
\end{equation}

This metric jointly captures both localization accuracy and classification performance.

\subsection{Baseline Methods and Comparative Experiments}

This section aims to validate the effectiveness of the proposed PASTE comprehensively. To ensure broad generalizability, we select baseline detectors that span single-stage and two-stage paradigms, CNN-based and Transformer-based backbones, as well as anchor-based and anchor-free designs. The selected detectors and their main characteristics are summarized in Table \ref{tab:baselines}, along with the rationale for their selection.

\begin{table*}[t]
\centering
\caption{Performance Boost by PASTE on FAIR-CSAR (SL Mode)}
\label{tab:sl-boost}
\small
\renewcommand{\arraystretch}{1.3}
\begin{tabular}{@{}
  >{\raggedright\arraybackslash}m{0.28\linewidth}
  >{\centering\arraybackslash}m{0.16\linewidth}
  >{\centering\arraybackslash}m{0.18\linewidth}
  >{\centering\arraybackslash}m{0.12\linewidth}
  >{\centering\arraybackslash}m{0.14\linewidth}@{}}
\toprule
\textbf{Baseline Detector} & \textbf{mAP@0.5} & \textbf{mAP@0.5 (+PASTE)} & \textbf{Gain} & \textbf{Relative gain (\%)} \\
\midrule
Rotated RetinaNet & 22.9 & 24 & $\mathbf{1.1\uparrow}$ & 4.8\% \\
GWD & 26 & 27 & $\mathbf{1.0\uparrow}$ & 3.8\% \\
RoI Transformer & 38 & 42.3 & $\mathbf{4.3\uparrow}$ & 11.3\% \\
Rotated FCOS & 34.5 & 35.7 & $\mathbf{1.2\uparrow}$ & 3.5\% \\
Oriented R-CNN (ResNet-50) & 39.2 & 43.3 & $\mathbf{4.1\uparrow}$ & 10.5\% \\
Oriented R-CNN (Swin-T) & 41.3 & 44.4 & $\mathbf{3.1\uparrow}$ & 7.5\% \\
\bottomrule
\end{tabular}
\end{table*}

\begin{table*}[t]
\centering
\caption{Performance Boost by PASTE on FAIR-CSAR (FSI Mode)}
\label{tab:fsi-boost}
\small
\renewcommand{\arraystretch}{1.3}
\begin{tabular}{@{}
  >{\raggedright\arraybackslash}m{0.28\linewidth}
  >{\centering\arraybackslash}m{0.16\linewidth}
  >{\centering\arraybackslash}m{0.18\linewidth}
  >{\centering\arraybackslash}m{0.12\linewidth}
  >{\centering\arraybackslash}m{0.14\linewidth}@{}}
\toprule
\textbf{Baseline Detector} & \textbf{mAP@0.5} & \textbf{mAP@0.5 (+PASTE)} & \textbf{Gain} & \textbf{Relative gain (\%)} \\
\midrule
Rotated RetinaNet & 37.2 & 38.9 & $\mathbf{1.7\uparrow}$ & 4.6\% \\
GWD & 39.4 & 40.6 & $\mathbf{1.2\uparrow}$ & 3.0\% \\
RoI Transformer & 45.4 & 48.7 & $\mathbf{3.3\uparrow}$ & 7.3\% \\
Rotated FCOS & 45.4 & 46.7 & $\mathbf{1.3\uparrow}$ & 2.9\% \\
Oriented R-CNN (ResNet-50) & 44.8 & 48.3 & $\mathbf{3.5\uparrow}$ & 7.8\% \\
Oriented R-CNN (Swin-T) & 50 & 52 & $\mathbf{2.0\uparrow}$ & 4.0\% \\
\bottomrule
\end{tabular}
\end{table*}

\begin{figure*}[t]
    \centering
    \includegraphics[width=1\linewidth]{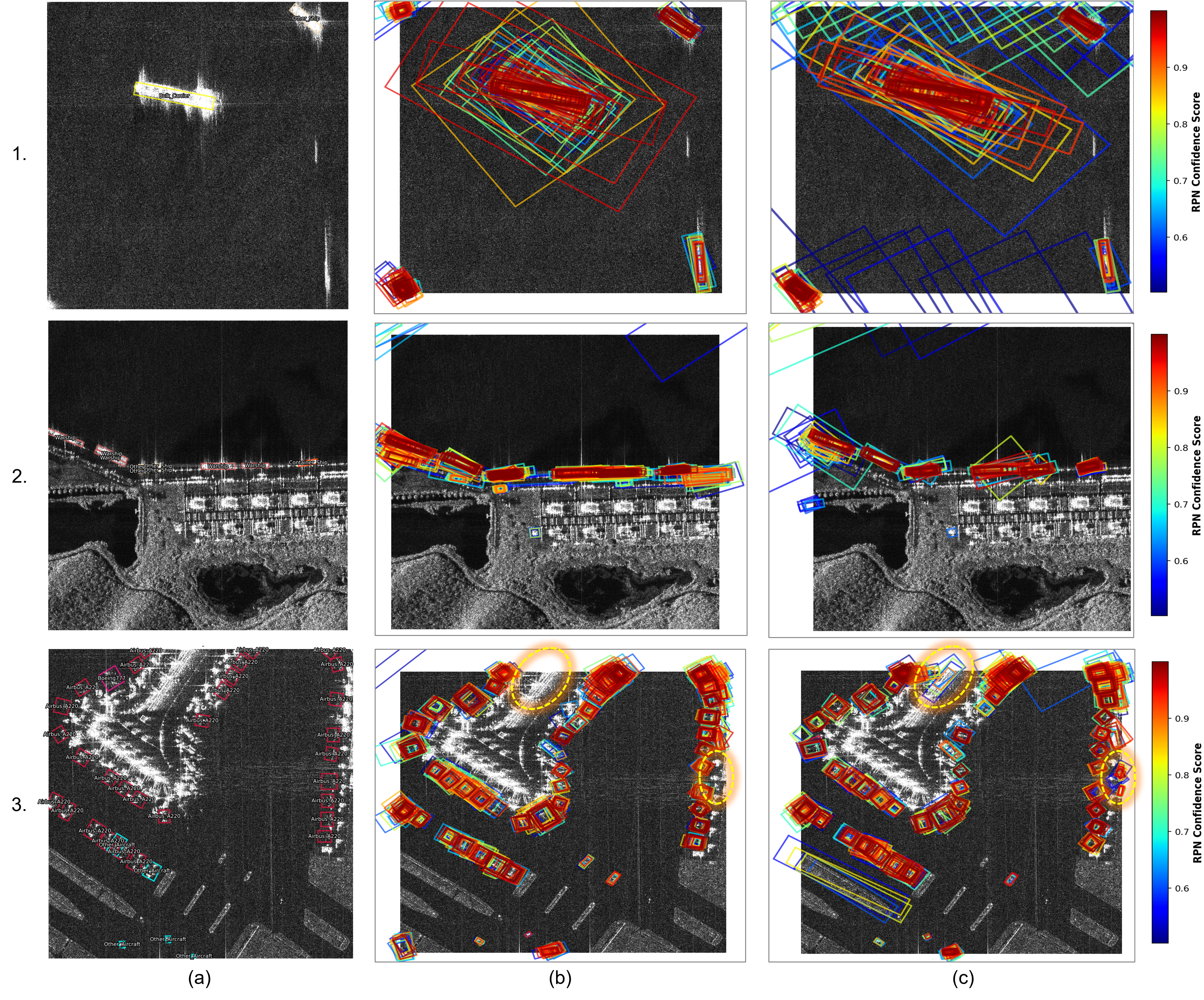}
    \caption{Proposals (First stage output) visualization under three scenes. Row 1: inshore ships; Row 2: offshore ships; Row 3: terminal aircraft. Column (a): Ground-truth boxes; Column (b): Top-scoring proposals ($\ge$ 0.5) generated by Oriented R-CNN equipped with PASTE; Column (c): Top-scoring proposals ($\ge$ 0.5) generated by Oriented R-CNN without PASTE. All proposals are colored from cool to warm according to confidence score.}
    \label{fig: vis_RPN}
\end{figure*}

All baseline detectors in Table. \ref{tab:baselines} were trained following the configurations described in Section \RNum{4}-B, and their best mAP@0.5 scores were recorded. Subsequently, each detector was re-trained under identical conditions after integrating the proposed PASTE module, and the corresponding optimal mAP@0.5 values were again logged. The performance gains conferred by PASTE under both SL and FSI modes on FAIR-CSAR are compactly reported in Table. \ref{tab:sl-boost} and Table. \ref{tab:fsi-boost}, respectively. 

As evidenced in Table \ref{tab:sl-boost} and \ref{tab:fsi-boost}, PASTE consistently yields performance improvements across all baseline detectors, with mAP@0.5 gains ranging from 1.0 to 4.3 percentage points. Notably, the most pronounced enhancement is observed on RoI Transformer (+4.3 pp), which underscores PASTE's efficacy in geometry-aware refinement. More importantly, for the strong Oriented R-CNN baseline, integrating PASTE into the ResNet-50 variant surpasses the performance gain achieved by upgrading the backbone to Swin-T ($\Delta$mAP = 4.1 pp vs. 2.1 pp). This head-to-head comparison validates that PASTE's contribution is both significant and complementary to architectural advances in feature extraction, thereby establishing its practical merit for SAR object detection.

Notably, PASTE yields substantially higher performance gains on two-stage detectors ($\Delta$mAP = 2.0-4.3 pp) compared with their single-stage counterparts ($\Delta$mAP = 1.0–1.2 pp). Two-stage detectors uniquely possess a proposal-generation stage in which the Region Proposal Network (RPN) produces object candidates. A plausible reason is that PASTE improves the accuracy of proposal generation. We therefore adopted the strong Oriented R-CNN baseline, loading two checkpoints—one trained with the PASTE module and the other without—and performed inference on a single test image. Subsequently, both qualitative and quantitative analyses were conducted to further evaluate the effect of PASTE on proposal generation.

\subsubsection{Qualitative Analysis of Impact on Proposal Generation by PASTE}

To qualitatively verify the hypothesis that PASTE enhances the quality of proposal generation, we visualized the RPN proposals together with the corresponding ground-truth (GT) boxes, as shown in Fig. \ref{fig: vis_RPN}. Three representative scenes were deliberately selected—offshore ships, inshore ships, and terminal aircraft—and arranged from top to bottom, respectively. From left to right, the three columns illustrate the GT boxes, the RPN proposals generated with PASTE, and those generated without PASTE. This visualization enables a direct comparison of spatial proposal distributions, providing intuitive evidence of the improvements introduced by PASTE.

Examining the first two rows (ship scenes), it can be observed that proposals generated with PASTE exhibit a high degree of shape consistency: their orientations and aspect ratios closely match and tightly cluster around the ground-truth boxes. In contrast, proposals produced without PASTE are markedly more scattered, with divergent orientations and aspect ratios even at the same location. This evidence that PASTE injects scattering-topology geometric cues, which impose an effective geometric constraint on the first-stage RPN. 

In terminal areas, aircraft are often parked flush against jet-bridges and buildings, making detection extremely challenging, as illustrated in the third row. Comparing the two visualizations reveals that the RPN without PASTE falsely outputs two high-confidence proposals inside the strongly scattering zones of the terminal building (highlighted by yellow dashed circles). Conversely, the PASTE-enhanced RPN produces no proposals in these regions, effectively filtering out potential false-alarm areas before they reach the second-stage detector. This further implies that PASTE endows the detector with a certain scattering-topology prior, enhancing its ability to discriminate between the topological signatures of terminal buildings and aircraft. A closer look at the apron area in the third row further reveals that the PASTE-equipped RPN tends to emit more proposals for small objects. This offers the second-stage detector a richer pool of candidates and thus boosts recall for diminutive targets; nevertheless, it also inevitably increases the risk of false positives among these tiny instances.

\begin{figure*}[t]
    \centering
    \includegraphics[width=1\linewidth]{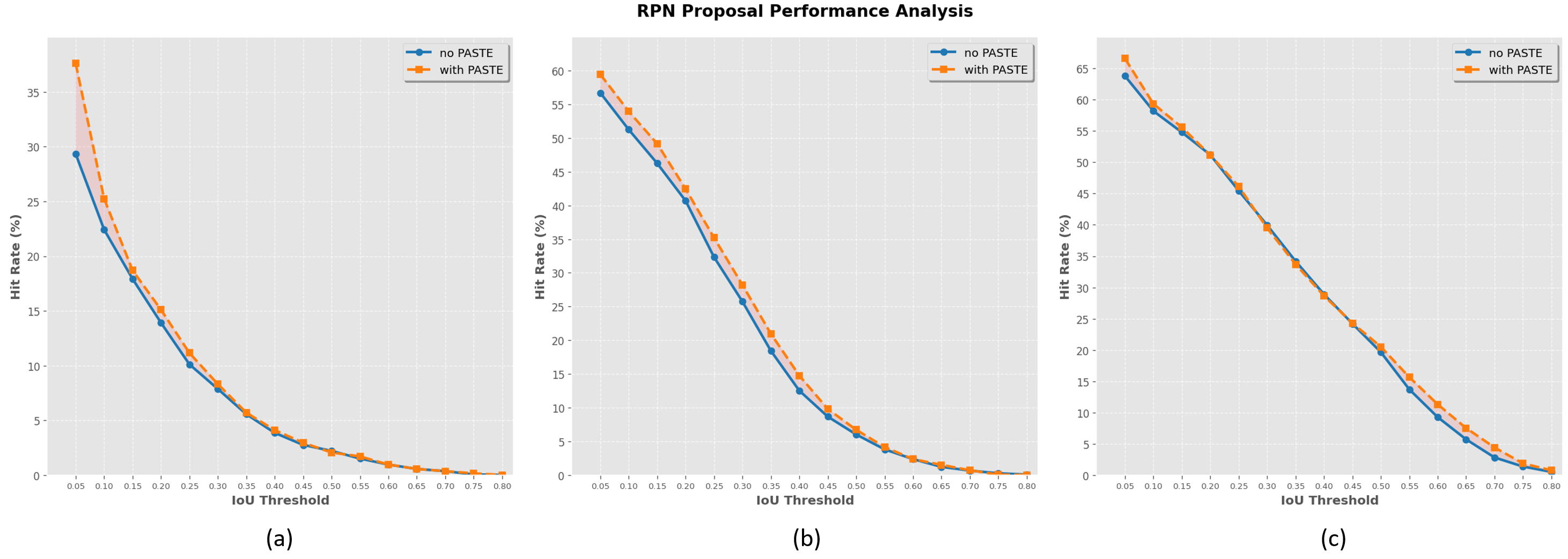}
    \caption{Hit Rate of predicted Proposals under variant IoU threshold under three typical detection scenes: (a) Offshore ships; (b) Inshore Ships; (c) Terminal aircrafts. }
    \label{fig: PHR_IOU}
\end{figure*}

\subsubsection{Quantitative Analysis of Impact on Proposal Generation by PASTE}

Following the qualitative visualization analysis, a quantitative evaluation of the impact of the PASTE on RPN-predicted proposals is further conducted across the three representative scenarios. Specifically, we begin by calculating the Proposal Hit Rate (PHR) between all generated proposals and the GT boxes across a range of IoU thresholds from 0.05 to 0.8, generating the curve shown in Fig. \ref{fig: PHR_IOU}. The PHR here is defined as the ratio of the number of proposals having an IoU greater than the given threshold with any ground truth box to the total number of proposals. In the ship detection scenario, the algorithm with the PASTE detector achieves a higher ground-truth hit rate for medium- and low-quality proposals with IoU confidence below 0.5, while the hit rate for high-quality proposals remains comparable to the baseline. In the complex airport terminal scene for aircraft detection, the algorithm with the PASTE detector achieves a higher ground-truth hit rate for proposals with IoU confidence below 0.15 and above 0.5, whereas for other IoU thresholds, the performance is generally consistent with the baseline. These results suggest that the PASTE module effectively enhances the discriminative capability of the detector when the proposal quality is limited, thereby improving the utilization efficiency of suboptimal region proposals in the two-stage detection framework. In contrast, for dense small targets such as aircraft, PASTE improves the hit rate at both extremely low and high IoU intervals, indicating its stronger ability to suppress feature confusion under complex background interference.

\begin{table}[t]
\centering
\caption{Precision of proposals in different scenarios}
\label{tab:proposal_precision}
\renewcommand{\arraystretch}{1.3}
\small
\begin{tabular}{@{}
  >{\raggedright\arraybackslash}m{0.3\linewidth}
  >{\centering\arraybackslash}m{0.3\linewidth}
  >{\centering\arraybackslash}m{0.3\linewidth}@{}}
\toprule
\textbf{Scenario} & \textbf{Proposal Precision (no PASTE)} & \textbf{Proposal Precision (with PASTE)} \\ 
\midrule
Offshore Ships & 81.4\% & 93.1\% \\
Inshore Ships & 56.7\% & 59.5\% \\
Terminal Aircrafts & 63.8\% & 66.6\% \\
\bottomrule
\end{tabular}
\end{table}

Subsequently, the precision of the proposals in each scene was calculated and summarized in Table \ref{tab:proposal_precision}, which is defined as the ratio of proposals containing ground-truth targets to the total number of proposals. A higher precision indicates that a larger proportion of proposals can effectively cover the true targets. The statistical results demonstrate that the introduction of the PASTE module leads to a significant improvement in proposal precision across all scenarios. 

Both the qualitative and quantitative analyses presented in this section indicate that PASTE can substantially enhance the quality of proposal generation in the region proposal stage, thereby contributing to the overall superior performance of two-stage detectors compared with single-stage ones.

\begin{figure}[t]
    \centering
    \includegraphics[width=1\linewidth]{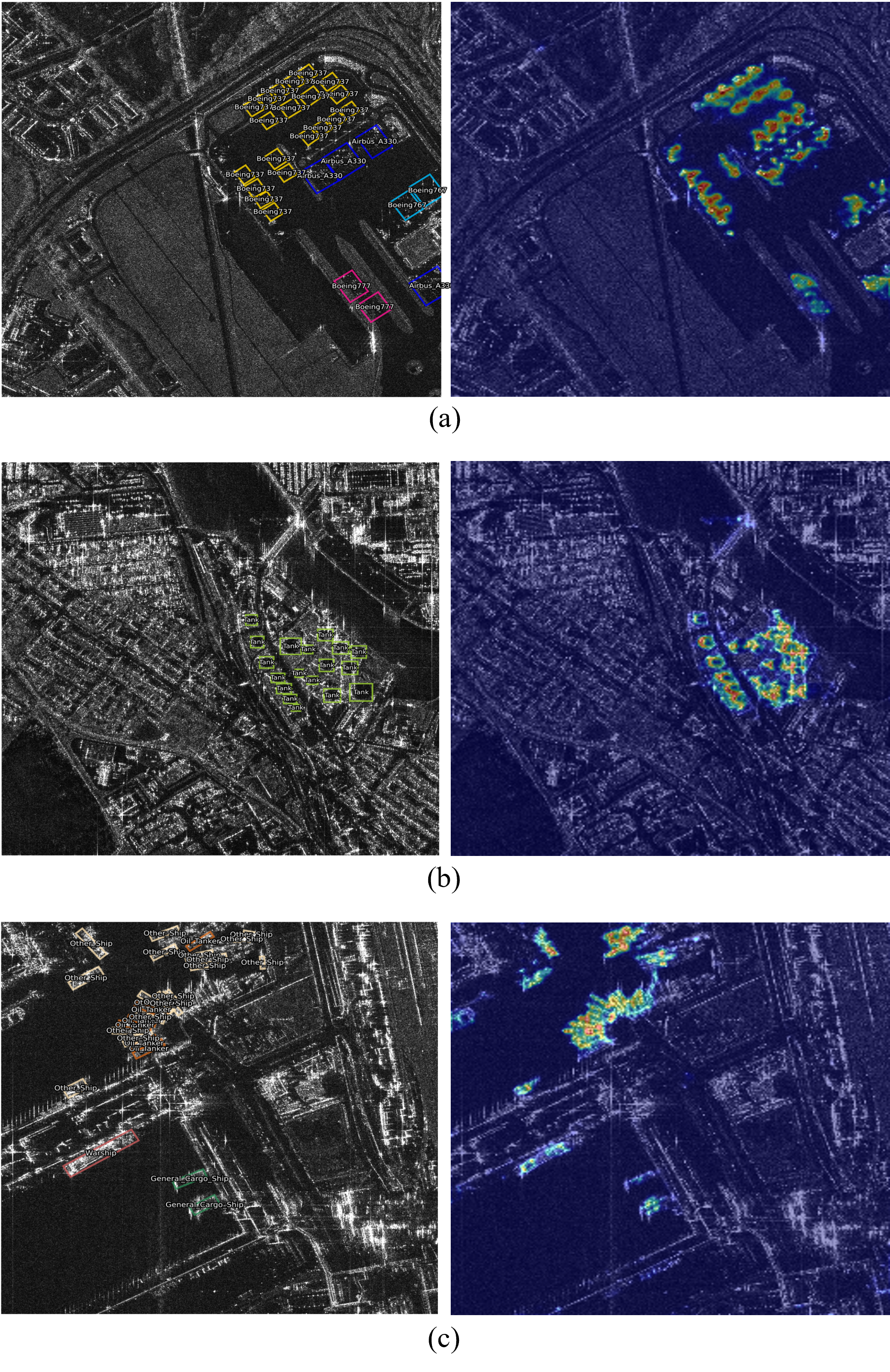}
    \caption{Visualization of predicted spatial scattering map in different scenarios. The left column shows the test samples with ground-truth bounding boxes, while the right column presents the corresponding spatial scattering map predicted by the proposed model. From top to bottom: (a) aircraft detection at an airport, (b) oil tank detection in a ground industrial area, and (c) near-shore ship detection at a port. The highlighted regions in the scattering map correspond to the areas where the prediction network in STIM focuses on the key scattering structures.}
    \label{fig:SAM vis}
\end{figure}

\begin{figure*}[t]
    \centering
    \includegraphics[width=\textwidth]{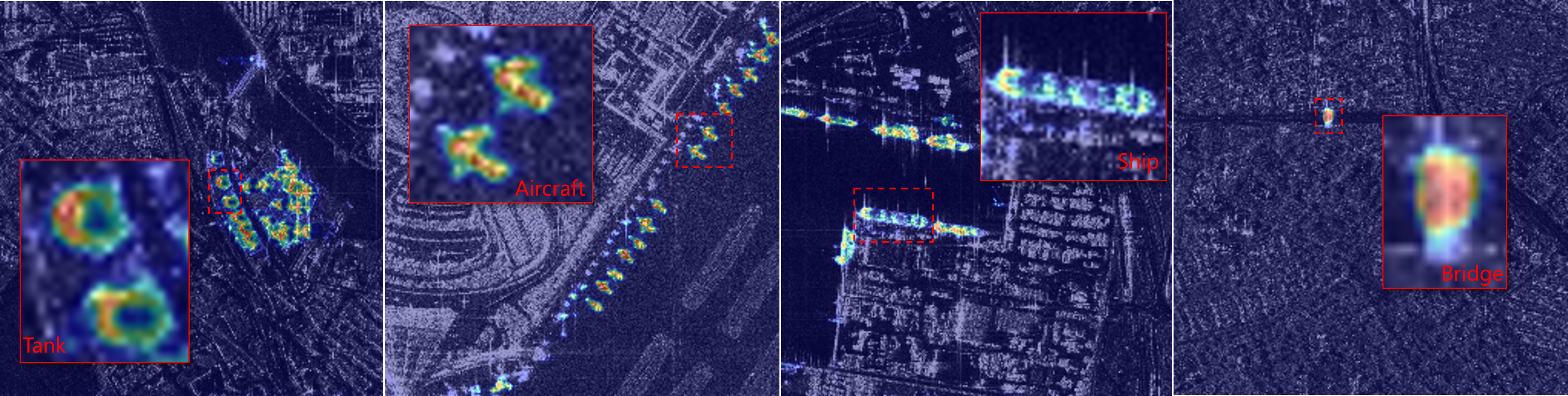}
    \caption{Predicted spatial scattering map on target region across representative categories.}
    \label{fig:msam_attention_visualization}
\end{figure*}

\subsection{Visual Analysis of Predicted Spatial Scattering Map}

To further interpret how the spatial scattering map predicted by the STIM contributes to the improvement of detection performance, Fig.~\ref{fig:SAM vis} visualizes the spatial attention prediction results in three representative scenarios. The left column shows the test samples with ground-truth bounding boxes, while the right column presents the corresponding spatial scattering map predicted by the proposed model.

Specifically, Fig.~\ref{fig:SAM vis}(a) corresponds to the airport scene, where the attention responses are mainly concentrated on the fuselage and wing regions of aircraft while effectively avoiding the strong scattering interference from terminal buildings. Fig.~\ref{fig:SAM vis}(b) shows the ground oil tank scene, in which the model’s attention is distributed along the tank boundaries, forming ring-shaped activation patterns that align well with the circular geometry of the storage tanks. Fig.~\ref{fig:SAM vis}(c) presents the near-shore port scene, where the attention is primarily focused on the ship contours and dock regions, while background clutter such as buildings and water surfaces exhibits weak activation. These visualization results indicate that the spatial attention prediction network, supervised by ASC-based keypoint annotations automatically generated through SKAA, can effectively capture the geometric and scattering structures of different types of targets in complex SAR scenes. The model not only accurately focuses on the main scattering regions of aircraft but also learns the circular topology of oil tanks and distinguishes foreground from background in complex near-shore environments. 

Furthermore, we analyze the predicted scattering maps within the target regions, as illustrated in Fig. \ref{fig:msam_attention_visualization}. Under the supervision of scattering keypoints, the attention prediction network achieves an almost perfect alignment with the scattering structures of the target regions, enabling effective discrimination between foreground and background. In some cases, the network can even explicitly reveal the geometric structure of the target. These observations suggest that, after training, the attention prediction branch inherently possesses latent capabilities for object detection and, to some extent, object segmentation.

Overall, these findings verify the effectiveness of SKAA-generated keypoint supervision and demonstrate the strong capability of the STIM module in predicting spatial attention consistent with the target scattering structures. This not only improves detection accuracy but also provides strong physical interpretability for the proposed fusion method.

\subsection{Ablation Studies of Main Components in PASTE}

In this subsection, three ablation studies are devised to dissect the contribution of each component in PASTE: the proposed STIM, SPSS, and the scattering keypoints automatically generated by the SKAA. Specifically, the ablation is organized into two sets according to the presence of the SPSS. In the set that retains SPSS, this part further compares two automatically generated keypoint sources: (i) scattering-aware keypoints produced by the proposed SKAA (hereafter termed SKAA-Keypoints), and (ii) amplitude-statistics-based local extrema (hereafter termed Amp-Keypoints). For the set trained without SPSS, no keypoint supervision is needed. 

To highlight the superiority of the proposed scattering keypoints derived from the ASC model's parameter inversion, a baseline
approach that extracts Amp-Keypoints directly from SAR images is introduced. Specifically, we employ a Difference-of-Gaussian (DoG)--based detector \cite{wang2013ship} to obtain salient amplitude extrema for comparison. Given an input amplitude image $I$, two
Gaussian-smoothed images are generated using standard deviations $\sigma_1 = 1.0$ and $\sigma_2 = 1.6$ with a $3 \times 3$ kernel. The DoG
response is then computed as
\begin{equation}
    D = G_{\sigma_2}(I) - G_{\sigma_1}(I).
\end{equation}
Spatial locations satisfying $|D(x,y)| > 5$ are identified as candidate keypoints, which are further ranked according to their absolute DoG magnitudes. The top-$30$ points are retained as amplitude-dominant keypoints. Then, these points are clustered into nine representative
keypoints for each target. This Amp-Keypoints provides a strong baseline for evaluating the advantages of the proposed ASC-based scattering keypoint generation method SKAA.

Owing to its learnable spatial-transformation layer and rotation-equivariant feature extraction, RoI Transformer exhibits intrinsically higher sensitivity to geometric-topology information; furthermore, it recorded the most pronounced performance gain when equipped with PASTE in Section \RNum{4}-D. Considering these characteristics,  RoI Transformer is adopted as the benchmark architecture for the ablation study. The corresponding results are summarized in Table \ref{tab: Ablation}. It is observed that the introduction of the STIM alone, when supervised by the downstream object detection loss function, yields only a limited performance improvement. Subsequently, a joint optimization incorporating the SPSS and STIM significantly enhances the model's performance, owing to the introduced keypoint topological prior. Specifically, the AMP-Keypoints generated by the DoG algorithm achieved an improvement of only 0.7\%, while the SKAA-Keypoints, derived from the ASC model, brought a more substantial performance gain of 2.4\%. This result strongly demonstrates that, for remote sensing target detection and recognition interpretation tasks, keypoints constructed from position parameters obtained via an electromagnetic parameterized representation model are more effective and possess stronger representational capabilities than those relying solely on image amplitude information.

\begin{figure}[t]
    \centering
    \includegraphics[width=1\linewidth]{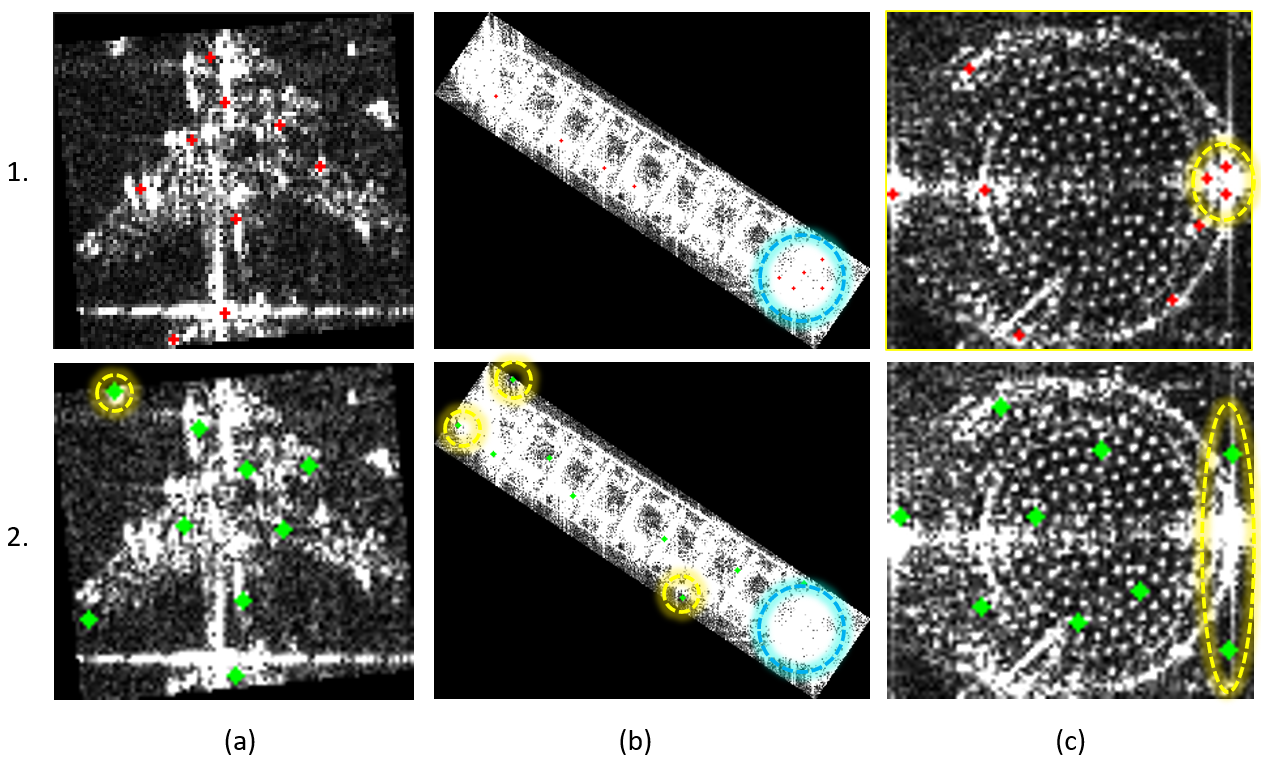}
    \caption{Visual examples of SKAA-Keypoints and AMP-Keypoints. Row 1: SKAA-Keypoints (Red); Row 2: AMP-Keypoints (Green); Column (a): Aircraft; Column (b): Ship; Column (c): Tank. }
    \label{fig: vis_Keypoints}
\end{figure}

\begin{table}[t]
    \centering
    \small
    \caption{Ablation studies for different components of PASTE.}
    \label{tab: Ablation}
    \renewcommand{\arraystretch}{1.3}
    \begin{tabular}{@{}
  >{\centering\arraybackslash}m{0.1\linewidth}
  >{\centering\arraybackslash}m{0.2\linewidth}
  >{\centering\arraybackslash}m{0.35\linewidth}
  >{\centering\arraybackslash}m{0.15\linewidth}@{}}
        \toprule
        \textbf{STIM} & \textbf{SPSS} & \textbf{Keypoints Type} & \textbf{mAP@0.5} \\
        \midrule
        \ding{55} & $\backslash$ & $\backslash$ & 38 \\
        \ding{51} & \ding{55} & $\backslash$ & 39.9 \\
        \ding{51} & \ding{51} & AMP-Keypoints & 40.6 \\
        \ding{51} & \ding{51} & SKAA-Keypoints & \textbf{42.3} \\
        \bottomrule
    \end{tabular}
\end{table}

Furthermore, to visually elucidate the superior scattering topological representation capability of the proposed SKAA-Keypoints, Fig. \ref{fig: vis_Keypoints} presents a comparison of the distribution differences between the two keypoint types. We select representative targets with distinct scattering characteristics, including an aircraft, a bulk carrier, and an oil tank. For the aircraft target, SKAA-Keypoints accurately localize the main scattering components—such as the nose, fuselage, engines, and tail—demonstrating robust structural consistency. In contrast, the amplitude-based AMP-Keypoints occasionally misidentify surrounding strong scatterers, for instance, reflections from ground service facilities near the aircraft, resulting in false responses outside the true target area (as indicated by the yellow dashed ellipse). For the ship target, the bridge region (highlighted by the blue dashed ellipse) contains dense and complex scattering components. The SKAA-Keypoints are coherently concentrated on the continuous strong-scattering structures along the bridge, forming a stable, linear, and geometrically consistent distribution aligned with the ship’s main axis. Conversely, the AMP-Keypoints are more dispersed and susceptible to high-energy sidelobe interference, causing some clustered points to deviate beyond the actual ship structure (as marked by the yellow dashed ellipse). In the tank region, SKAA-Keypoints remain tightly clustered over the main scattering body of the target, maintaining strong structural consistency. In comparison, the amplitude-based AMP-Keypoints are prone to sidelobe contamination, generating several false detections outside the true target region (as marked by the yellow dashed ellipse).

\subsection{Ablation Studies of Keypoint Number and SPSS Hyper-parameters}

\begin{table}[t]
  \centering
  \caption{Ablation Study on Different Numbers of Keypoints}
  \label{tab:keypoint_map}
  \small
  \renewcommand{\arraystretch}{1.2}  % row spacing
\begin{tabular}{@{}
>{\centering\arraybackslash}m{0.4\linewidth}
>{\centering\arraybackslash}m{0.3\linewidth}@{}}
    \toprule
    \textbf{Number of Keypoints} & \textbf{mAP@0.5} \\
    \midrule
    3  & 41.6 \\
    6  & 41.4 \\
    \textbf{9}  & \textbf{43.3} \\
    12 & 42.6 \\
    15 & 42.1 \\
    \bottomrule
  \end{tabular}
\end{table}

% In the preamble:
% \usepackage{booktabs}
% \usepackage{array}

\begin{table}[!t]
\centering
\caption{Ablation Studies on SPSS Hyperparameters}
\label{tab:sga-ablation}
\small
\renewcommand{\arraystretch}{1.2}
\setlength{\tabcolsep}{16pt}
\begin{tabular}{
    c c
    c c
}
\toprule
\multicolumn{2}{c}{\textbf{SPSS Weight $\lambda$}} &
\multicolumn{2}{c}{\textbf{Gaussian Radius $\sigma$}} \\
\cmidrule(lr){1-2} \cmidrule(lr){3-4}
\textbf{$\lambda$} & \textbf{mAP@0.5} &
\textbf{$\sigma$}   & \textbf{mAP@0.5} \\
\midrule
0.1 & 42.5 & \textbf{1} & \textbf{43.3} \\
0.5 & 42.2 & 2 & 42.4 \\
\textbf{1} & \textbf{43.3} & 4 & 42.1 \\
2 & 41.8 & 8 & 41.1 \\
\bottomrule
\end{tabular}
\end{table}

\begin{table*}[t]
\small
\centering
\caption{Computational cost and memory overhead of proposed PASTE.}
\begin{threeparttable}
\setlength{\tabcolsep}{20pt}
\renewcommand{\arraystretch}{1.3}
\begin{tabular}{l l c c c}
\toprule
\textbf{Stage} & \textbf{Metric} & \textbf{Baseline} & \textbf{PASTE} & \textbf{Overhead} \\ 
\midrule

\multirow{1}{*}{\textbf{Annotation Generation}} 
    & Avg time per sample (ms) & -- & 413.5 & -- \\ 
\midrule

\multirow{2}{*}{\textbf{Training}} 
    & Model size (MB) & 315.2 & 337.7 & +7.1\% \\
    & Training time per epoch (s) & 588.4 & 661.5 & +12\% \\
\midrule

\multirow{2}{*}{\textbf{Inference}} 
    & FLOPs (G) & 211.44 & 274.3 & +29.7\% \\
    & Latency per image (ms) & 61.7 & 72.4 & +17.3\% \\
\bottomrule
\end{tabular}
\begin{tablenotes}
\footnotesize
\item Overhead is calculated relative to the baseline detector. 
\end{tablenotes}
\end{threeparttable}
\label{tab:ASTFM_overhead}
\end{table*}

To investigate the influence of the key hyperparameters on the detection performance, we conducted ablation studies on two critical factors of the proposed method: (1) the number of keypoints generated by SKAA, and (2) the hyperparameters in the SPSS, i.e., the SPSS weight $\lambda$ and the Gaussian radius $\sigma$. The Oriented R-CNN is selected as the baseline detector for performance evaluation in this subsection.

\subsubsection{Effect of Keypoint Number}

Table~\ref{tab:keypoint_map} summarizes the detection performance with different numbers of keypoints. It can be observed that the mAP first increases as the number of keypoints grows and then slightly decreases when the number exceeds a certain threshold. Specifically, the model achieves the best performance (43.3\% mAP@0.5) when using nine keypoints, indicating that an appropriate number of keypoints effectively balances spatial coverage and feature redundancy. Using too few keypoints (e.g., three or six) may result in insufficient characterization of the target topology and limited spatial attention, whereas using too many keypoints (e.g., twelve or fifteen) may introduce noisy regions and redundant features, thereby weakening the prominence of the primary spatial information.

\subsubsection{Effect of SPSS Hyper-parameters}

The results of varying $\lambda$ and $\sigma$ are presented in Table~\ref{tab:sga-ablation}. The ablation study on the SPSS weight was conducted while keeping the weights of the bounding box regression loss and the classification loss fixed at 1. We evaluated the detection performance of models trained with different SPSS weights $\lambda = 0.1, 0.5, 1,$ and $2$. The results indicate that the detector achieves the best mAP when $\lambda = 1$, where the SPSS and the detection losses (classification and regression) contribute equally, accounting for approximately 50\% each.

For the Gaussian radius $\sigma$, we tested values of 1, 2, 4, and 8. The results show that the detection performance gradually decreases as $\sigma$ increases, dropping from an mAP of 43.3\% to lower values. This trend suggests that the spatial attention distribution must be precisely aligned with the scattering structures of the targets. When the Gaussian radius becomes too large, the attention becomes overly diffused across scattered keypoints, which introduces blurred geometric structures and weakens the constraint imposed by keypoint information. Consequently, the enhancement effect of the scattering keypoints diminishes.

\subsection{Spatiotemporal Resource Consumption Analysis}

This section analyzes the computational cost and memory overhead introduced by the proposed method. First, the runtime of the SKAA process is evaluated by measuring the time required to generate scattering keypoint annotations for all training samples in a detection dataset, which reflects the practicality and efficiency of the method. Subsequently, the additional storage requirements and training time incurred by incorporating scattering keypoint annotations are examined. Finally, the computational and memory overhead introduced during the inference phase is also assessed. Oriented R-CNN is selected as the baseline detector for testing.

As the results illustrated in Table \ref{tab:ASTFM_overhead}, the proposed PASTE introduces only a modest computational and memory overhead compared to the baseline detector. Firstly, at the annotation generation stage, the proposed SKAA process exhibits high efficiency in the annotation stage. It takes only an average of 413.5 ms to process a single training sample. This efficient design enables the entire FAIR-CSAR dataset, which contains over 340K instances, to be automatically annotated with adaptive scattering keypoints within one day. Such efficiency substantially reduces the manual annotation workload and ensures the scalability of the proposed PASTE framework to large-scale SAR target detection tasks. In the training phase, the model size increases from 315.2 MB to 337.7 MB, corresponding to a 7.1\% growth. The training time per epoch rises by 12\%, from 588.4 s to 661.5 s, indicating that the adaptive structure incurs limited training complexity. During inference, due to the additional scattering map prediction network in STIM, the number of FLOPs increases by 29.7\% (from 211.44 G to 274.3 G), and the latency per image increases by 17.3\% (from 61.7 ms to 72.4 ms). Despite these increases, the computational overhead remains acceptable given the performance gains achieved by PASTE in detection accuracy and robustness.

\section{Conclusion}

This paper presented the PASTE to address the fundamental limitations of directly transferring optical-domain detection paradigms to SAR imagery. Specifically, PASTE is built upon three tightly coupled components. First, the proposed SKAA method enables scalable, zero-manual-cost generation of scattering-center–level annotations by inverting ASC position parameters within existing target-level labels, providing physically consistent and topology-aware supervision. Second, the STIM injects the learned scattering topology into multi-scale feature representations through spatial scattering maps, guiding the detector to prioritize stable electromagnetic structures over noisy texture patterns. Third, the SPSS introduces a Gaussian-based soft supervision mechanism that aligns network optimization with the spatial distribution of scattering centers, thereby enforcing a physically meaningful attention pattern during training without compromising end-to-end inference efficiency.

Extensive experiments conducted on the FAIR-CSAR benchmark demonstrate that PASTE is architecture-agnostic and can be seamlessly integrated into a wide range of mainstream detectors, including single-stage, two-stage, anchor-based, anchor-free, and Transformer-based frameworks. Across all evaluated baselines, PASTE consistently yields significant performance improvements, with relative mAP gains ranging from 2.9\% to 11.3\%. Detailed ablation studies further confirm that ASC-derived scattering keypoints substantially outperform amplitude-based keypoints, highlighting the superiority of physics-driven scattering topology as a robust structural point for detection and recognition in SAR scenes. In addition, visual analysis shows that the scatter attention map predicted by the PASTE framework has the ability to distinguish between foreground and background, which well demonstrates the network's understanding of the target scattering topology structure and demonstrates interpretability.

% {\appendix[Proof of the Zonklar Equations]
% Use $\backslash${\tt{appendix}} if you have a single appendix:
% Do not use $\backslash${\tt{section}} anymore after $\backslash${\tt{appendix}}, only $\backslash${\tt{section*}}.
% If you have multiple appendixes use $\backslash${\tt{appendices}} then use $\backslash${\tt{section}} to start each appendix.
% You must declare a $\backslash${\tt{section}} before using any $\backslash${\tt{subsection}} or using $\backslash${\tt{label}} ($\backslash${\tt{appendices}} by itself
%  starts a section numbered zero.)}


\bibliographystyle{IEEEtran}
\bibliography{ref}

\begin{thebibliography}{10}
\providecommand{\url}[1]{#1}
\csname url@samestyle\endcsname
\providecommand{\newblock}{\relax}
\providecommand{\bibinfo}[2]{#2}
\providecommand{\BIBentrySTDinterwordspacing}{\spaceskip=0pt\relax}
\providecommand{\BIBentryALTinterwordstretchfactor}{4}
\providecommand{\BIBentryALTinterwordspacing}{\spaceskip=\fontdimen2\font plus
\BIBentryALTinterwordstretchfactor\fontdimen3\font minus \fontdimen4\font\relax}
\providecommand{\BIBforeignlanguage}[2]{{%
\expandafter\ifx\csname l@#1\endcsname\relax
\typeout{** WARNING: IEEEtran.bst: No hyphenation pattern has been}%
\typeout{** loaded for the language `#1'. Using the pattern for}%
\typeout{** the default language instead.}%
\else
\language=\csname l@#1\endcsname
\fi
#2}}
\providecommand{\BIBdecl}{\relax}
\BIBdecl

\bibitem{survey1}
J.~Slesinski and D.~Wierzbicki, ``Review of synthetic aperture radar automatic target recognition: A dual perspective on classical and deep learning techniques,'' \emph{IEEE Journal of Selected Topics in Applied Earth Observations and Remote Sensing}, vol.~18, pp. 18\,978--19\,024, 2025.

\bibitem{survey2}
X.~X. Zhu, D.~Tuia, L.~Mou, G.-S. Xia, L.~Zhang, F.~Xu, and F.~Fraundorfer, ``Deep learning in remote sensing: A comprehensive review and list of resources,'' \emph{IEEE Geoscience and Remote Sensing Magazine}, vol.~5, no.~4, pp. 8--36, 2017.

\bibitem{zhang2023remote}
X.~Zhang, T.~Zhang, G.~Wang, P.~Zhu, X.~Tang, X.~Jia, and L.~Jiao, ``Remote sensing object detection meets deep learning: A metareview of challenges and advances,'' \emph{IEEE Geoscience and Remote Sensing Magazine}, vol.~11, no.~4, pp. 8--44, 2023.

\bibitem{HUGHES2020166}
\BIBentryALTinterwordspacing
L.~H. Hughes, D.~Marcos, S.~Lobry, D.~Tuia, and M.~Schmitt, ``A deep learning framework for matching of sar and optical imagery,'' \emph{ISPRS Journal of Photogrammetry and Remote Sensing}, vol. 169, pp. 166--179, 2020. [Online]. Available: \url{https://www.sciencedirect.com/science/article/pii/S0924271620302598}
\BIBentrySTDinterwordspacing

\bibitem{li2021multiscale}
Y.~Li, L.~Du, and D.~Wei, ``Multiscale cnn based on component analysis for sar atr,'' \emph{IEEE Transactions on Geoscience and Remote Sensing}, vol.~60, pp. 1--12, 2021.

\bibitem{li2019sar}
T.~Li and L.~Du, ``Sar automatic target recognition based on attribute scattering center model and discriminative dictionary learning,'' \emph{IEEE Sensors Journal}, vol.~19, no.~12, pp. 4598--4611, 2019.

\bibitem{Hu2024Conceptual}
F.~Hu, F.~Xu, R.~Wang, X.~Qiu, C.~Ding, and Y.~Jin, ``Conceptual study and performance analysis of tandem multi-antenna spaceborne sar interferometry,'' \emph{Journal of Remote Sensing}, vol.~4, 2024.

\bibitem{Guo2020ExplainableSAR}
W.~Guo, Z.~Zhang, W.~Yu \emph{et~al.}, ``{Perspective on Explainable SAR Target Recognition},'' \emph{Journal of Radars}, vol.~9, no.~3, pp. 462--476, 2020.

\bibitem{HZL2020}
Z.~Huang, Z.~Pan, and B.~Lei, ``What, where, and how to transfer in sar target recognition based on deep cnns,'' \emph{IEEE Transactions on Geoscience and Remote Sensing}, vol.~58, no.~4, pp. 2324--2336, 2020.

\bibitem{feng2022electromagnetic}
S.~Feng, K.~Ji, F.~Wang, L.~Zhang, X.~Ma, and G.~Kuang, ``Electromagnetic scattering feature (esf) module embedded network based on asc model for robust and interpretable sar atr,'' \emph{IEEE Transactions on Geoscience and Remote Sensing}, vol.~60, pp. 1--15, 2022.

\bibitem{zhang2020fec}
J.~Zhang, M.~Xing, and Y.~Xie, ``Fec: A feature fusion framework for sar target recognition based on electromagnetic scattering features and deep cnn features,'' \emph{IEEE Transactions on Geoscience and Remote Sensing}, vol.~59, no.~3, pp. 2174--2187, 2020.

\bibitem{GNN-FiLM}
J.~Hou, Z.~Bian, G.~Yao, H.~Lin, Y.~Zhang, S.~He, and H.~Chen, ``Attribute scattering center-assisted sar atr based on gnn-film,'' \emph{IEEE Geoscience and Remote Sensing Letters}, vol.~21, pp. 1--5, 2024.

\bibitem{zhang2021integrating}
J.~Zhang, M.~Xing, G.-C. Sun, and Z.~Bao, ``Integrating the reconstructed scattering center feature maps with deep cnn feature maps for automatic sar target recognition,'' \emph{IEEE Geoscience and Remote Sensing Letters}, vol.~19, pp. 1--5, 2021.

\bibitem{liu2021eftl}
J.~Liu, M.~Xing, H.~Yu, and G.~Sun, ``Eftl: Complex convolutional networks with electromagnetic feature transfer learning for sar target recognition,'' \emph{IEEE Transactions on Geoscience and Remote Sensing}, vol.~60, pp. 1--11, 2021.

\bibitem{ren2015faster}
S.~Ren, K.~He, R.~Girshick, and J.~Sun, ``Faster r-cnn: Towards real-time object detection with region proposal networks,'' \emph{Advances in neural information processing systems}, vol.~28, 2015.

\bibitem{cai2018cascade}
Z.~Cai and N.~Vasconcelos, ``Cascade r-cnn: Delving into high quality object detection,'' in \emph{Proceedings of the IEEE conference on computer vision and pattern recognition}, 2018, pp. 6154--6162.

\bibitem{xie2021oriented}
X.~Xie, G.~Cheng, J.~Wang, X.~Yao, and J.~Han, ``Oriented r-cnn for object detection,'' in \emph{Proceedings of the IEEE/CVF international conference on computer vision}, 2021, pp. 3520--3529.

\bibitem{yang2021r3det}
X.~Yang, J.~Yan, Z.~Feng, and T.~He, ``R3det: Refined single-stage detector with feature refinement for rotating object,'' in \emph{Proceedings of the AAAI conference on artificial intelligence}, vol.~35, no.~4, 2021, pp. 3163--3171.

\bibitem{han2021align}
J.~Han, J.~Ding, J.~Li, and G.-S. Xia, ``Align deep features for oriented object detection,'' \emph{IEEE transactions on geoscience and remote sensing}, vol.~60, pp. 1--11, 2021.

\bibitem{tian2019fcos}
Z.~Tian, C.~Shen, H.~Chen, and T.~He, ``Fcos: Fully convolutional one-stage object detection,'' in \emph{Proceedings of the IEEE/CVF international conference on computer vision}, 2019, pp. 9627--9636.

\bibitem{duan2019centernet}
K.~Duan, S.~Bai, L.~Xie, H.~Qi, Q.~Huang, and Q.~Tian, ``Centernet: Keypoint triplets for object detection,'' in \emph{Proceedings of the IEEE/CVF international conference on computer vision}, 2019, pp. 6569--6578.

\bibitem{redmon2016you}
J.~Redmon, S.~Divvala, R.~Girshick, and A.~Farhadi, ``You only look once: Unified, real-time object detection,'' in \emph{Proceedings of the IEEE conference on computer vision and pattern recognition}, 2016, pp. 779--788.

\bibitem{cai2024yolov8}
Y.~Cai, Y.~Zhao, S.~Wen, and J.~Feng, ``Improved yolov8 sar image aircraft object detection method,'' in \emph{2024 7th International Symposium on Autonomous Systems (ISAS)}, 2024, pp. 1--6.

\bibitem{carion2020end}
N.~Carion, F.~Massa, G.~Synnaeve, N.~Usunier, A.~Kirillov, and S.~Zagoruyko, ``End-to-end object detection with transformers,'' in \emph{European conference on computer vision}.\hskip 1em plus 0.5em minus 0.4em\relax Springer, 2020, pp. 213--229.

\bibitem{2024OEGR}
Y.~Feng, Y.~You, J.~Tian, and G.~Meng, ``Oegr-detr: A novel detection transformer based on orientation enhancement and group relations for sar object detection,'' \emph{Remote Sensing}, vol.~16, no.~1, p.~28, 2024.

\bibitem{yin2024review}
J.~Yin, C.~Duan, H.~Wang, and J.~Yang, ``A review on the few-shot sar target recognition,'' \emph{IEEE Journal of Selected Topics in Applied Earth Observations and Remote Sensing}, 2024.

\bibitem{zhou2024simulated}
X.~Zhou, T.~Tang, Q.~He, L.~Zhao, G.~Kuang, and L.~Liu, ``Simulated sar prior knowledge guided evidential deep learning for reliable few-shot sar target recognition,'' \emph{ISPRS Journal of Photogrammetry and Remote Sensing}, vol. 216, pp. 1--14, 2024.

\bibitem{wang2021multichannel}
Y.~Wang, J.~Cheng, Y.~Zhou, F.~Zhang, and Q.~Yin, ``A multichannel fusion convolutional neural network based on scattering mechanism for polsar image classification,'' \emph{IEEE Geoscience and Remote Sensing Letters}, vol.~19, pp. 1--5, 2021.

\bibitem{makhija2024polsar}
S.~Makhija, S.~Mandal, U.~Pandya, S.~Chirakkal, and D.~Putrevu, ``Polsar image classification using complex-valued squeeze and excitation network,'' in \emph{International Conference on Pattern Recognition}.\hskip 1em plus 0.5em minus 0.4em\relax Springer, 2024, pp. 270--286.

\bibitem{shi2025multi}
J.~Shi, S.~Ji, H.~Jin, Y.~Zhang, M.~Gong, and W.~Lin, ``Multi-feature lightweight deeplabv3+ network for polarimetric sar image classification with attention mechanism,'' \emph{Remote Sensing}, vol.~17, no.~8, p. 1422, 2025.

\bibitem{imani2025attention}
M.~Imani, ``Attention based network for fusion of polarimetric and contextual features for polarimetric synthetic aperture radar image classification,'' \emph{Engineering Applications of Artificial Intelligence}, vol. 139, p. 109665, 2025.

\bibitem{wang2023target}
R.~Wang, Z.~Wang, Y.~Chen, H.~Kang, F.~Luo, and Y.~Liu, ``Target recognition in sar images using complex-valued network guided with sub-aperture decomposition,'' \emph{Remote Sensing}, vol.~15, no.~16, p. 4031, 2023.

\bibitem{wu2024fair}
Y.~Wu, Y.~Suo, Q.~Meng, W.~Dai, T.~Miao, W.~Zhao, Z.~Yan, W.~Diao, G.~Xie, Q.~Ke \emph{et~al.}, ``Fair-csar: A benchmark dataset for fine-grained object detection and recognition based on single look complex sar images,'' \emph{IEEE Transactions on Geoscience and Remote Sensing}, 2024.

\bibitem{ozkaya2020automatic}
U.~{\"O}zkaya, ``Automatic target recognition (atr) from sar imaginary by using machine learning techniques,'' \emph{Avrupa Bilim ve Teknoloji Dergisi}, pp. 165--169, 2020.

\bibitem{shao2023CFARG}
Z.~Shao, X.~Zhang, X.~Xu, T.~Zeng, T.~Zhang, and J.~Shi, ``Cfar-guided convolution neural network for large scale scene sar ship detection,'' in \emph{2023 IEEE Radar Conference (RadarConf23)}, 2023, pp. 1--5.

\bibitem{hu2019aircraft}
H.~Hu, L.~Huang, and W.~Yu, ``Aircraft detection for hr sar images in non-homogeneous background using ggmd-based modeling,'' \emph{Chinese Journal of Electronics}, vol.~28, no.~6, pp. 1271--1280, 2019.

\bibitem{ding2017robust}
B.~Ding, G.~Wen, J.~Zhong, C.~Ma, and X.~Yang, ``A robust similarity measure for attributed scattering center sets with application to sar atr,'' \emph{Neurocomputing}, vol. 219, pp. 130--143, 2017.

\bibitem{potter1997attributed}
L.~C. Potter and R.~L. Moses, ``Attributed scattering centers for sar atr,'' \emph{IEEE Transactions on image processing}, vol.~6, no.~1, pp. 79--91, 1997.

\bibitem{chen2014CNN}
S.~Chen and H.~Wang, ``Sar target recognition based on deep learning,'' in \emph{2014 International Conference on Data Science and Advanced Analytics (DSAA)}, 2014, pp. 541--547.

\bibitem{TNN_cui}
J.~Cui, H.~Jia, H.~Wang, and F.~Xu, ``A fast threshold neural network for ship detection in large-scene sar images,'' \emph{IEEE Journal of Selected Topics in Applied Earth Observations and Remote Sensing}, vol.~15, pp. 6016--6032, 2022.

\bibitem{chen2024yolo}
J.~Chen, Y.~Shen, Y.~Liang, Z.~Wang, and Q.~Zhang, ``Yolo-sad: An efficient sar aircraft detection network,'' \emph{Applied Sciences}, vol.~14, no.~7, p. 3025, 2024.

\bibitem{NIPS2017_Transformer}
\BIBentryALTinterwordspacing
A.~Vaswani, N.~Shazeer, N.~Parmar, J.~Uszkoreit, L.~Jones, A.~N. Gomez, L.~u. Kaiser, and I.~Polosukhin, ``Attention is all you need,'' in \emph{Advances in Neural Information Processing Systems}, I.~Guyon, U.~V. Luxburg, S.~Bengio, H.~Wallach, R.~Fergus, S.~Vishwanathan, and R.~Garnett, Eds., vol.~30.\hskip 1em plus 0.5em minus 0.4em\relax Curran Associates, Inc., 2017. [Online]. Available: \url{https://proceedings.neurips.cc/paper_files/paper/2017/file/3f5ee243547dee91fbd053c1c4a845aa-Paper.pdf}
\BIBentrySTDinterwordspacing

\bibitem{chen2022geospatial}
L.~Chen, R.~Luo, J.~Xing, Z.~Li, Z.~Yuan, and X.~Cai, ``Geospatial transformer is what you need for aircraft detection in sar imagery,'' \emph{IEEE Transactions on Geoscience and Remote Sensing}, vol.~60, pp. 1--15, 2022.

\bibitem{Jia2024Ship}
H.~Jia, X.~Pu, Q.~Liu, H.~Wang, and F.~Xu, ``A fast progressive ship detection method for very large full-scene sar images,'' \emph{IEEE Transactions on Geoscience and Remote Sensing}, vol.~62, pp. 1--15, 2024.

\bibitem{fusarShip2020}
\BIBentryALTinterwordspacing
X.~Hou, W.~Ao, Q.~Song, J.~Lai, H.~Wang, and F.~Xu, ``Fusar-ship: building a high-resolution sar-ais matchup dataset of gaofen-3 for ship detection and recognition,'' \emph{Sci. China Inf. Sci.}, vol.~63, no.~4, 2020. [Online]. Available: \url{https://doi.org/10.1007/s11432-019-2772-5}
\BIBentrySTDinterwordspacing

\bibitem{10.1093/nsr/nwae403}
\BIBentryALTinterwordspacing
S.~Yin, C.~Fu, S.~Zhao, K.~Li, X.~Sun, T.~Xu, and E.~Chen, ``A survey on multimodal large language models,'' \emph{National Science Review}, vol.~11, no.~12, p. nwae403, 11 2024. [Online]. Available: \url{https://doi.org/10.1093/nsr/nwae403}
\BIBentrySTDinterwordspacing

\bibitem{MA2026VLM}
\BIBentryALTinterwordspacing
X.~Ma, H.~Xie, and S.~J. Qin, ``Efficiently integrate large language models with visual perception: A survey from the training paradigm perspective,'' \emph{Information Fusion}, vol. 125, p. 103419, 2026. [Online]. Available: \url{https://www.sciencedirect.com/science/article/pii/S1566253525004920}
\BIBentrySTDinterwordspacing

\bibitem{Kang2022SFR}
Y.~Kang, Z.~Wang, J.~Fu, X.~Sun, and K.~Fu, ``Sfr-net: Scattering feature relation network for aircraft detection in complex sar images,'' \emph{IEEE Transactions on Geoscience and Remote Sensing}, vol.~60, pp. 1--17, 2022.

\bibitem{Meng2025STC}
Q.~Meng, Y.~Wu, Y.~Suo, W.~Dai, Z.~Yan, X.~Gao, W.~Diao, and X.~Sun, ``Stc-net: Scattering topology cue-based network for aircraft detection in sar images,'' \emph{IEEE Transactions on Geoscience and Remote Sensing}, vol.~63, pp. 1--16, 2025.

\bibitem{xiao2025RPL}
X.~Xiao, Z.~Li, R.~Zhang, J.~Chen, and H.~Wang, ``Reciprocal point learning network with large electromagnetic kernel for sar open-set recognition,'' \emph{IEEE Transactions on Aerospace and Electronic Systems}, pp. 1--16, 2025.

\bibitem{fu2021scattering}
K.~Fu, J.~Fu, Z.~Wang, and X.~Sun, ``Scattering-keypoint-guided network for oriented ship detection in high-resolution and large-scale sar images,'' \emph{IEEE Journal of Selected Topics in Applied Earth Observations and Remote Sensing}, vol.~14, pp. 11\,162--11\,178, 2021.

\bibitem{guo2020scattering}
Q.~Guo, H.~Wang, and F.~Xu, ``Scattering enhanced attention pyramid network for aircraft detection in sar images,'' \emph{IEEE Transactions on Geoscience and Remote Sensing}, vol.~59, no.~9, pp. 7570--7587, 2020.

\bibitem{sun2022span}
Y.~Sun, Z.~Wang, X.~Sun, and K.~Fu, ``Span: Strong scattering point aware network for ship detection and classification in large-scale sar imagery,'' \emph{IEEE Journal of Selected Topics in Applied Earth Observations and Remote Sensing}, vol.~15, pp. 1188--1204, 2022.

\bibitem{wang2025attributed}
D.~Wang, Y.~Song, L.~Chen, and D.~An, ``Attributed scattering center guided network based on omnidirectional sub-aperture division for sar target detection,'' \emph{IEEE Transactions on Geoscience and Remote Sensing}, 2025.

\bibitem{pan2024sffnet}
X.~Pan, M.~Han, G.~Liao, L.~Yang, R.~Shao, and Y.~Li, ``Sffnet: A ship detection method using scattering feature fusion for sea surface sar images,'' \emph{IEEE Geoscience and Remote Sensing Letters}, 2024.

\bibitem{YY2025SARDet}
Y.~Yang, Z.~Lei, X.~Mo, D.~Lu, H.~Jia, and H.~Wang, ``Sardet-cl: Self-supervised contrastive learning with feature enhancement and imaging mechanism constraints for sar target detection,'' \emph{IEEE Transactions on Geoscience and Remote Sensing}, vol.~63, pp. 1--15, 2025.

\bibitem{Zhao2024Azimuth}
Y.~Zhao, L.~Zhao, S.~Zhang, K.~Ji, G.~Kuang, and L.~Liu, ``Azimuth-aware subspace classifier for few-shot class-incremental sar atr,'' \emph{IEEE Transactions on Geoscience and Remote Sensing}, vol.~62, pp. 1--20, 2024.

\bibitem{Huang2025PGD}
Z.~Huang, L.~Liu, S.~Yang, Z.~Wang, G.~Cheng, and J.~Han, ``Physics-guided detector for sar airplanes,'' \emph{IEEE Transactions on Circuits and Systems for Video Technology}, vol.~35, no.~12, pp. 12\,082--12\,095, 2025.

\bibitem{Chen2024RL_SAR}
J.~Chen, X.~Zhang, H.~Wang, and F.~Xu, ``A reinforcement learning framework for scattering feature extraction and sar image interpretation,'' \emph{IEEE Transactions on Geoscience and Remote Sensing}, vol.~62, pp. 1--14, 2024.

\bibitem{Zhang2009ASC}
A.~Zhang, ``Attributed scattering center feature extraction of complex target from high resolution sar imagery,'' Master's thesis, National University of Defense Technology, 2009.

\bibitem{wang2013ship}
Z.~Wang, C.~Wang, F.~Wu, B.~Zhang, H.~Zhang, and Y.~Tang, ``Ship detection for radarsat-2 scansar data using dog scale-space,'' in \emph{2013 IEEE International Geoscience and Remote Sensing Symposium-IGARSS}.\hskip 1em plus 0.5em minus 0.4em\relax IEEE, 2013, pp. 1881--1884.

\end{thebibliography}

\vfill

\end{document}